\documentclass[11pt]{article}
\usepackage{booktabs}       
\usepackage{amsfonts}       
\usepackage{amsmath}
\usepackage{graphicx}
\usepackage{amsthm}
\usepackage{amssymb}
\usepackage{multirow}
\usepackage{makecell}
\usepackage[vlined,linesnumbered,ruled,resetcount]{algorithm2e}
\usepackage[round]{natbib}
\usepackage[colorlinks,linkcolor=magenta,filecolor=blue,citecolor=blue,urlcolor=blue]{hyperref}%
\usepackage[top=1in, left=1in, right=1in, bottom=1in]{geometry}
\usepackage{subfigure}
\usepackage{amssymb}
\usepackage{pifont}
\usepackage{mathtools}
\usepackage{stmaryrd}
\usepackage[T1]{fontenc}

\begin{document}

\title{pGMM Kernel Regression and Comparisons with Boosted Trees}

\author{\vspace{0.5in}\\\textbf{Ping Li} and \textbf{Weijie Zhao}\\\\
Cognitive Computing Lab\\
Baidu Research\\
10900 NE 8th St. Bellevue, WA 98004, USA\\\\
\texttt{\{pingli98, zhaoweijie12\}@gmail.com}
}

\date{June 2022}

\maketitle
\begin{abstract}\vspace{0.3in}
\noindent In this work, we demonstrate the  advantage of the pGMM (``powered generalized min-max'') kernel in the context of (ridge) regression. In recent prior studies, the pGMM kernel has been extensively evaluated for classification tasks~\citep{li2017tunable,li2018several,li2022gcwsnet}, for logistic regression, support vector machines, as well as deep neural networks. In this paper, we provide an experimental study on ridge regression, to compare the pGMM kernel regression with the ordinary ridge linear regression as well as the RBF kernel ridge regression. Perhaps surprisingly, even without a tuning parameter (i.e., $p=1$ for the power parameter of the pGMM kernel), the pGMM kernel already performs  well. Furthermore, by tuning the  parameter $p$, this (deceptively simple) pGMM kernel even performs quite comparably to boosted trees.

\vspace{0.3in}

\noindent Boosting and boosted trees are very popular in machine learning practice. For regression tasks, typically, practitioners use $L_2$ boost, i.e., for minimizing the $L_2$ loss.  Sometimes for the purpose of robustness, the $L_1$ boost might be a choice. In this study, we implement $L_p$ boost for $p\geq 1$ and include it in the package of ``Fast ABC-Boost''~\citep{li2009abc,li2010robust,li2022fast}. Perhaps also surprisingly, the best performance (in terms of $L_2$ regression loss) is often attained at $p>2$, in some cases at $p\gg 2$. This phenomenon  has  already been demonstrated by~\citet{li2010approximating}  in the context of k-nearest neighbor classification using $L_p$ distances. In summary, the implementation of $L_p$ boost provides practitioners the additional flexibility of tuning boosting algorithms for potentially achieving better accuracy in regression applications.

\end{abstract}

\newpage\clearpage

\section{Introduction}

Kernels, which can be either linear or nonlinear, are classical tools in machine  learning~\citep{vapnik1999nature,scholkopf2002learning,bishop2006pattern,hastie2009elements}. In general, machine learning researchers and practitioners are often  familiar with the Gaussian (a.k.a. RBF, radial basis function) kernels, the polynomial kernels, etc. In this paper, we introduce the ``pGMM'' ($p$-powered generalized min-max) kernel for regression tasks.

The idea of the pGMM  kernel has been developed in a series of works over the recent years by~\cite{li2017tunable,li2018several,li2022gcwsnet}. To explain the pGMM kernel, we would need to first introduce the GMM (``generalized min-max'') kernel~\citep{li2017linearized,li2017theory}. The original ``min-max'' similarity (kernel) was  defined on non-negative data~\citep{kleinberg1999approximation,charikar2002similarity,manasse2010consistent,ioffe2010improved}. Thus,  in order to define the GMM kernel for general data types, the first step is to apply a transformation on the original data to remove negative entries via a simple dimensionality expansion.

Specifically, consider the original data vector $u_i$, $i=1$ to $d$. The following transformation, depending on whether an entry $u_i$ is positive or negative,
\begin{align}\label{eqn:transform}
 \left\{\begin{array}{cc}
\hspace{-.11in}\tilde{u}_{2i-1} = u_i,\hspace{0.1in} \tilde{u}_{2i} = 0&\text{if } \ u_i >0\\
\tilde{u}_{2i-1} = 0,\hspace{0.1in} \tilde{u}_{2i} =  -u_i &\text{if } \ u_i \leq 0
\end{array}\right.
\end{align}
converts general data types to non-negative data. Take a concrete example here. When $d=3$ and $u = [-3\ \ 17\ \ -0.8]$, the transformed data vector becomes $\tilde{u} = [0\ \ 3\ \ 17\ \ 0\ \ 0\ \ 0.8]$. The GMM kernel is then defined as the following form:
\begin{align}\label{eqn:GMM}
{\text{GMM}}(u,v) = \frac{\sum_{i=1}^{2d}\min\{\tilde{u}_i,\tilde{v}_i\}}{\sum_{i=1}^{2d} \max\{\tilde{u}_i,\tilde{v}_i\}}.
\end{align}
Note that, unlike the popular Gaussian (RBF) kernel,  the GMM kernel~\eqref{eqn:GMM} has no tuning parameter. To introduce tuning parameters,  the following pGMM kernel is one (deceptively) simple strategy:
\begin{align}\label{eqn:pGMM}
&{\text{pGMM}}(u,v;p) =  \frac{\sum_{i=1}^{2d}\left(\min\{\tilde{u}_i,\tilde{v}_i\}\right)^p}{\sum_{i=1}^{2d}\left( \max\{\tilde{u}_i,\tilde{v}_i\}\right)^p},
\end{align}
where $p\in\mathbb{R}$ is a tuning parameter. Readers have probably realized that this operation is  mathematically equivalent to applying a $p$-power transformation on the data before computing the GMM kernel. However, directly applying power transformations on the original data is typically not a  good idea due to numerical issues. As demonstrated in~\cite{li2017tunable,li2018several,li2022gcwsnet}, the pGMM kernel can be conveniently ``hashed'' using for example consistent weighted sampling (CWS)~\citep{manasse2010consistent,ioffe2010improved} or extremal process~\citep{li2021consistent}. Those hashing methods provide a numerically stable framework for applying the power transformation.

\vspace{0.1in}

As one can see, this pGMM kernel (and the related hashing algorithms) are constructed  from the original data without knowing the subsequent tasks, which can be  clustering, regression, or classification. As demonstrated in the previous papers or unpublished technical reports~\citep{li2017tunable,li2018several,li2022gcwsnet}, the pGMM kernel (with merely one tuning parameter) performs  surprisingly well in many classification tasks. It is expected that pGMM kernel should also work well for other tasks such as regression. Nevertheless, readers might be still interested in seeing those experiments. This motivates us to report the regression experiments using pGMM kernels.

\newpage

\noindent\textbf{Summary of contributions.} \ In this paper, we collect a set of regression datasets to compare pGMM kernel (ridge) regression with linear (ridge) regression and RBF kernel regression. Since the authors have been working on boosting and trees~\citep{li2009abc,li2010robust,li2022fast}, we also provide the $L_p$ regression  results on those datasets using boosted trees. The experimental results confirm the following:
\begin{itemize}
    \item The pGMM kernel performs (perhaps surprisingly) well on regression tasks.
    \item Boosted trees typically outperform the pGMM kernel, but usually not by too much.
    \item $L_p$ boost with $p> 2$ in many cases outperforms the standard $L_2$ (or $L_1$) boost.
\end{itemize}

\section{pGMM Kernel Ridge Regression}

Consider a training dataset: $X\in\mathbb{R}^{n\times d}$ and $Y\in\mathbb{R}^{n}$, where $n$ is the number of training examples and $d$ is the number of dimensions. Given a testing set $X_t$, the classical ridge linear regression (LR) outputs the prediction $\hat{Y}_t$ as
\begin{align}
\hat{Y}_t = X_t\left(X^TX+\lambda I_d\right)^{-1}X^TY,
\end{align}
where $I_d$ is the identity matrix of size $d\times d$ and $\lambda\geq 0$ is the regularization coefficient.

To help readers understand (pGMM) kernel regression, we make use of the following standard linear algebra trick (i.e., the ``push-through identity''):
\begin{align}
\left(X^TX+\lambda I_d\right)^{-1}X^T   = X^T\left(XX^T+\lambda I_n\right)^{-1}
\end{align}
to re-write ridge regression into a mathematically equivalent form:
\begin{align}
\hat{Y}_t = X_tX^T\left(XX^T+\lambda I_n\right)^{-1}Y,
\end{align}
or in a ``kernel'' form:
\begin{align}
\hat{Y}_t = K_t\left(K+\lambda I_n\right)^{-1}Y, \hspace{0.2in} K_t = X_tX^T,\hspace{0.2in} K = XX^T
\end{align}
The kernels $K = XX^T$ and $K_t = X_tX^T$ are simple linear kernels. One can of course also use nonlinear kernels, for example, the well-known Gaussian kernel or polynomial kernel. In this paper, we focus on the pGMM kernel
\begin{align}\label{eqn:KpGMM}
K(u,v;p) =     {\text{pGMM}}(u,v;p) =  \frac{\sum_{i=1}^{2d}\left(\min\{\tilde{u}_i,\tilde{v}_i\}\right)^p}{\sum_{i=1}^{2d}\left( \max\{\tilde{u}_i,\tilde{v}_i\}\right)^p}.
\end{align}
where $u$ and $v$ are two data vectors in the training set $X$ or testing set $X_t$.

Thus, in a most naive format, conducting kernel (ridge) regression can be extremely simple if the dataset size is small, e.g., with fewer than $30,000$ training examples, so that we can store the kernel matrix. This is of course not realistic (and not necessary) for larger datasets. The prior studies~\citep{li2017tunable,li2017linearized,li2017theory,li2018several,li2021consistent,li2022gcwsnet} have already provided very extensive experiments to demonstrate how to effectively linearize the min-max kernel and variants via hashing, for classification tasks. In this paper, for the sake of easy reproducibility, we just provide experiments directly using kernels, on relatively small datasets.

\newpage

\begin{table}[t]
\caption{Datasets for testing regression algorithms. We report the best test mean square error (MSE) for each method, over the range of regularization coefficients and parameters.}
\begin{center}{
\begin{tabular}{l r r r r r r r r}
\hline \hline
dataset & \# train & \# test &dim &LR&RBF&GMM&pGMM&$L_2$-Boost\\
\hline
ENBcool &384&384&8&10.24&3.20&1.70 &1.28 &1.21\\
ENBheat &384&384&8&9.00&0.495&0.191 &0.188 &0.186\\
Airfoil  & 752 & 751 & 5 &24.26&8.35 &7.50&3.56 &3.09\\
CPUsmall&4096&4096&12&102.12&9.05&7.22&7.05&6.89\\
CPU&4096&4096&21&98.35&6.42&5.17&5.03 &4.69\\
WECPerth &5000&5000&32&$1.1\times 10^9$&$2.3\times  10^8$&$2.4\times10^8$ &$2.3\times10^8$&$2.5\times10^8$\\
Cadata &10320&10320&8&$4.8\times10^9$&$3.8\times10^9$&$2.4\times10^9$&$2.4\times10^9$&$2.1\times10^9$\\
House16H &11392&11392&16&$2.1\times10^9$&$1.3\times10^9$&$1.2\times10^9$&$1.1\times10^9$&$1.0\times10^9$\\
House16L &11392&11392&16&$1.9\times 10^9$&$1.1\times10^9$&$9.9\times10^8$&$9.4\times10^8$&$8.6\times10^8$\\
CASP &22865&22865&9&26.63&23.94&15.30&15.29&13.51 \\\hline
Splice  & 1000& 2175  & 60 &0.1205& 0.0967 &0.0589 &0.0589 &0.0352\\
Mnoise1  & 2519& 454  & 784 &0.0484 &0.0344&0.0311 &0.0169&0.0145\\
Mnoise6  & 2519& 454  & 784 &0.0215& 0.0165 &0.0195&0.0129&0.0131\\
Mimage  & 2538& 10524 & 784 &0.0540 &0.0323 &0.0263&0.0125&0.0149\\
\hline\hline
\end{tabular}
}
\end{center}
\label{tab:data}
\end{table}

Next, we move directly to experiments. In Table~\ref{tab:data}, we  collect 14 datasets for regression from public sources, and report the test mean square  errors (MSEs) for ordinary ridge linear regression (LR), RBF kernel regression, GMM kernel regression, pGMM kernel regression, and $L_2$ boost.  Most of the datasets are from the UCI repository. The ``CPU'' and ``CPUsmall'' datasets are available at \url{http://www.cs.toronto.edu/~delve/data/datasets.html} .  In Table~\ref{tab:data}, the last three datasets (``Mnoise1'', ``Mnoise6'', and ``Mimage'') were created by~\citet{larochelle2007empirical} for testing deep learning algorithms by modifying the original MNIST (hand-written digits) dataset. Here, we only use the samples corresponding to digit ``0'' and digit ``1''. Note that those three datasets as well as the ``Splice'' dataset are originally for classification  tasks. We simply treat them as regression data for regressing on the \{0,1\} label values. In addition, we preprocess all datasets in the same fashion, by linearly scaling each feature to [0, 1]. Some kind of data preprocessing such as scaling is typically needed for kernel (similarity) based methods. Note that tree methods are invariant to monotonic scaling of each feature.

\begin{figure}[h]
\begin{center}

\mbox{
    \includegraphics[width=2.7in]{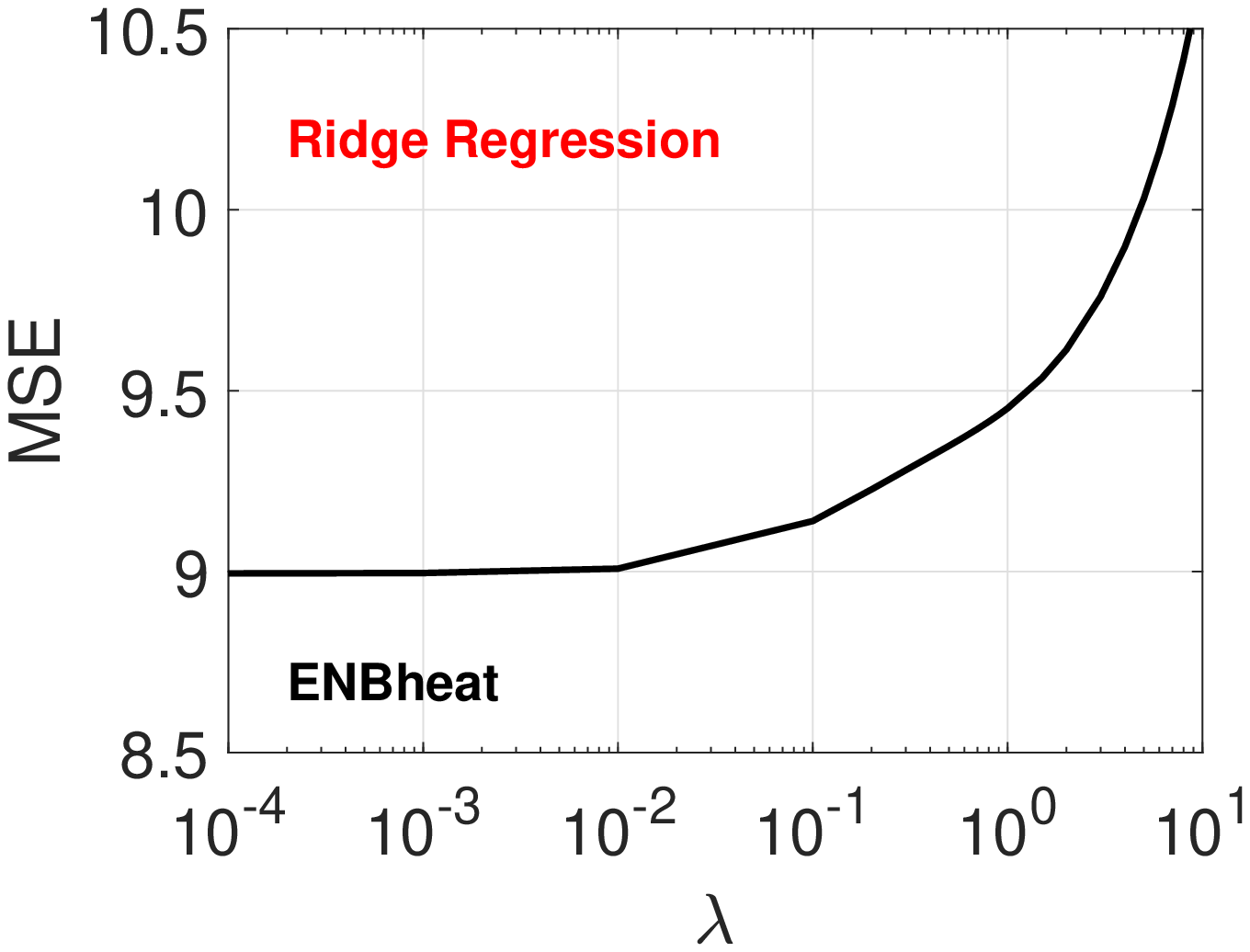}
    \includegraphics[width=2.7in]{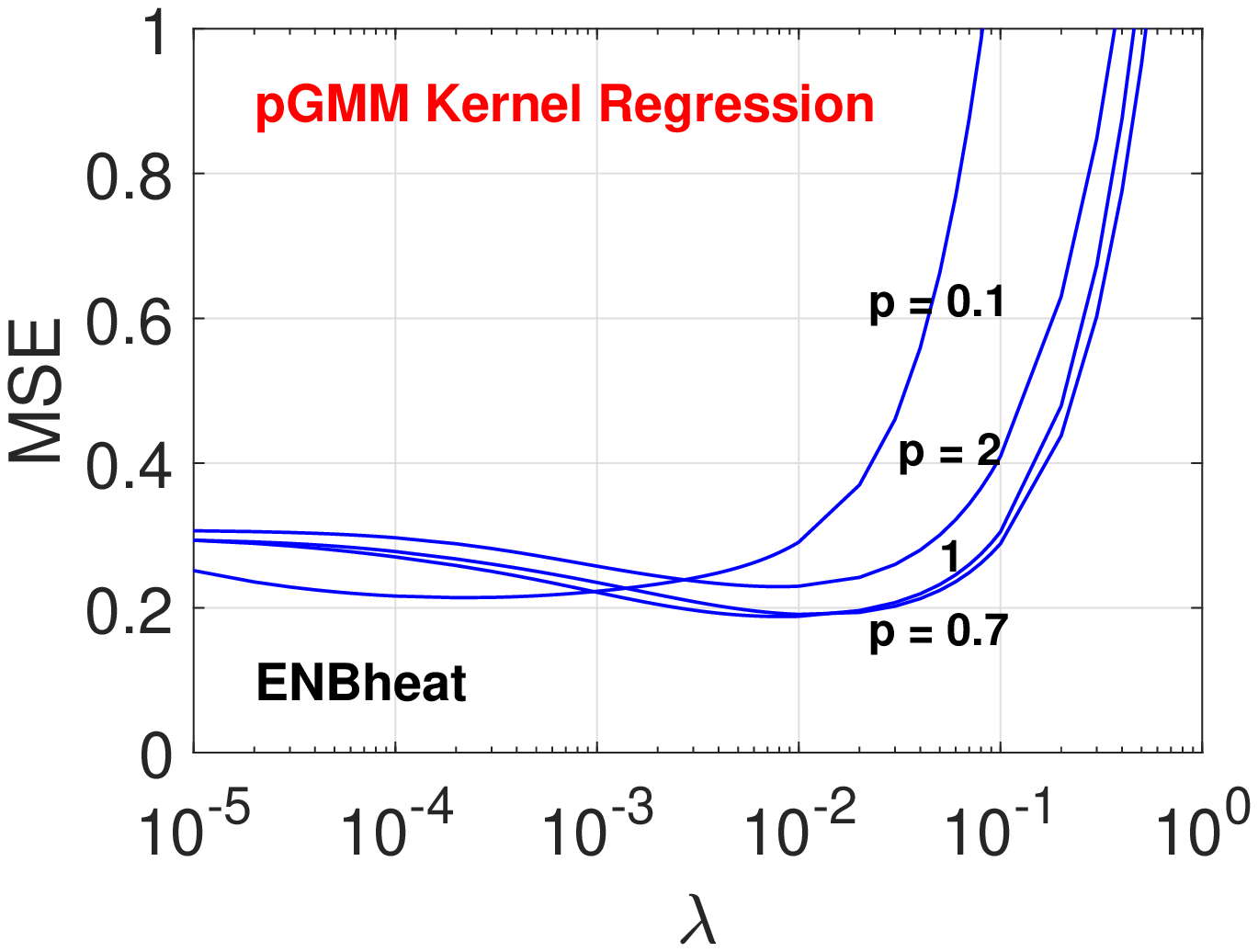}
}

\mbox{
    \includegraphics[width=2.7in]{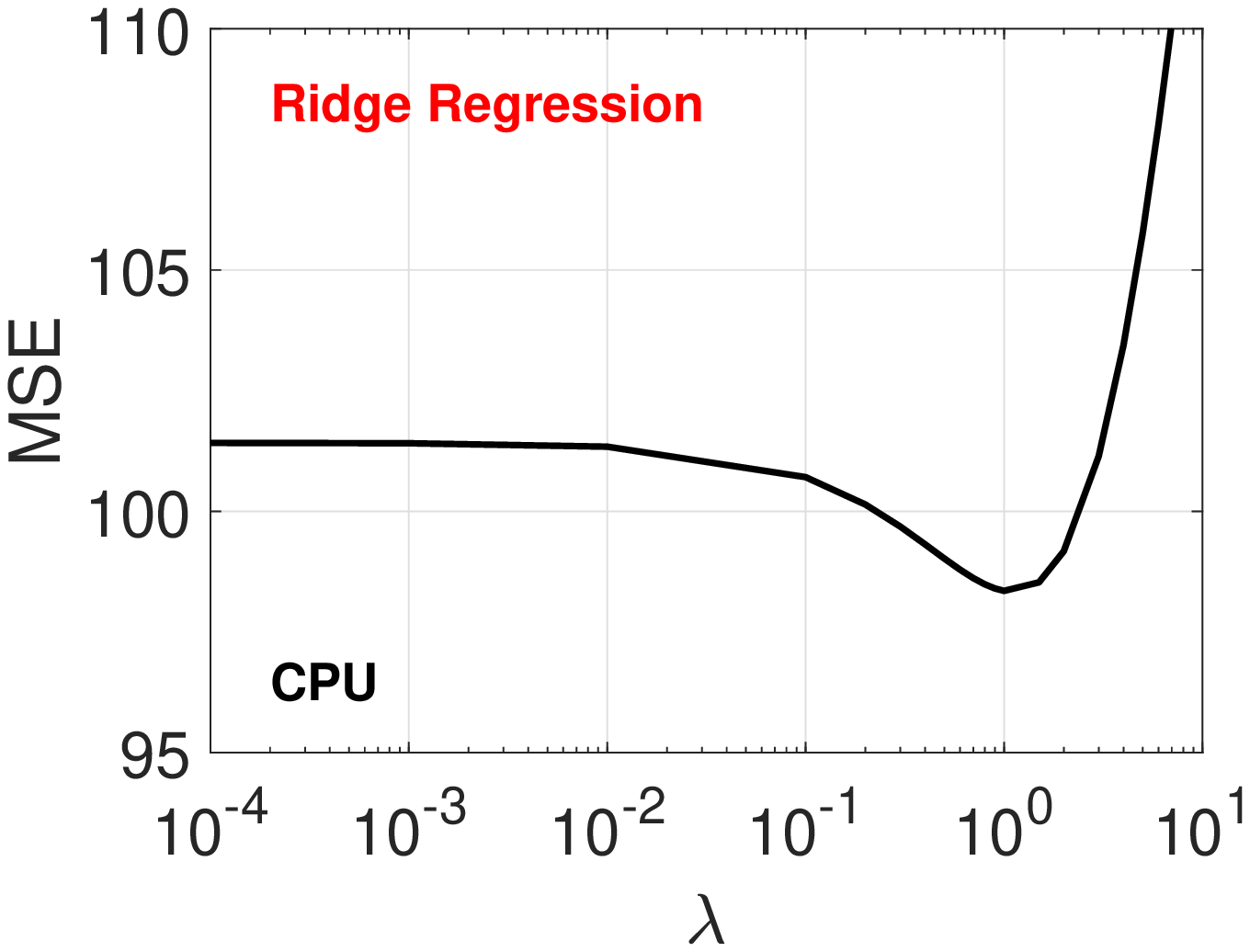}
    \includegraphics[width=2.7in]{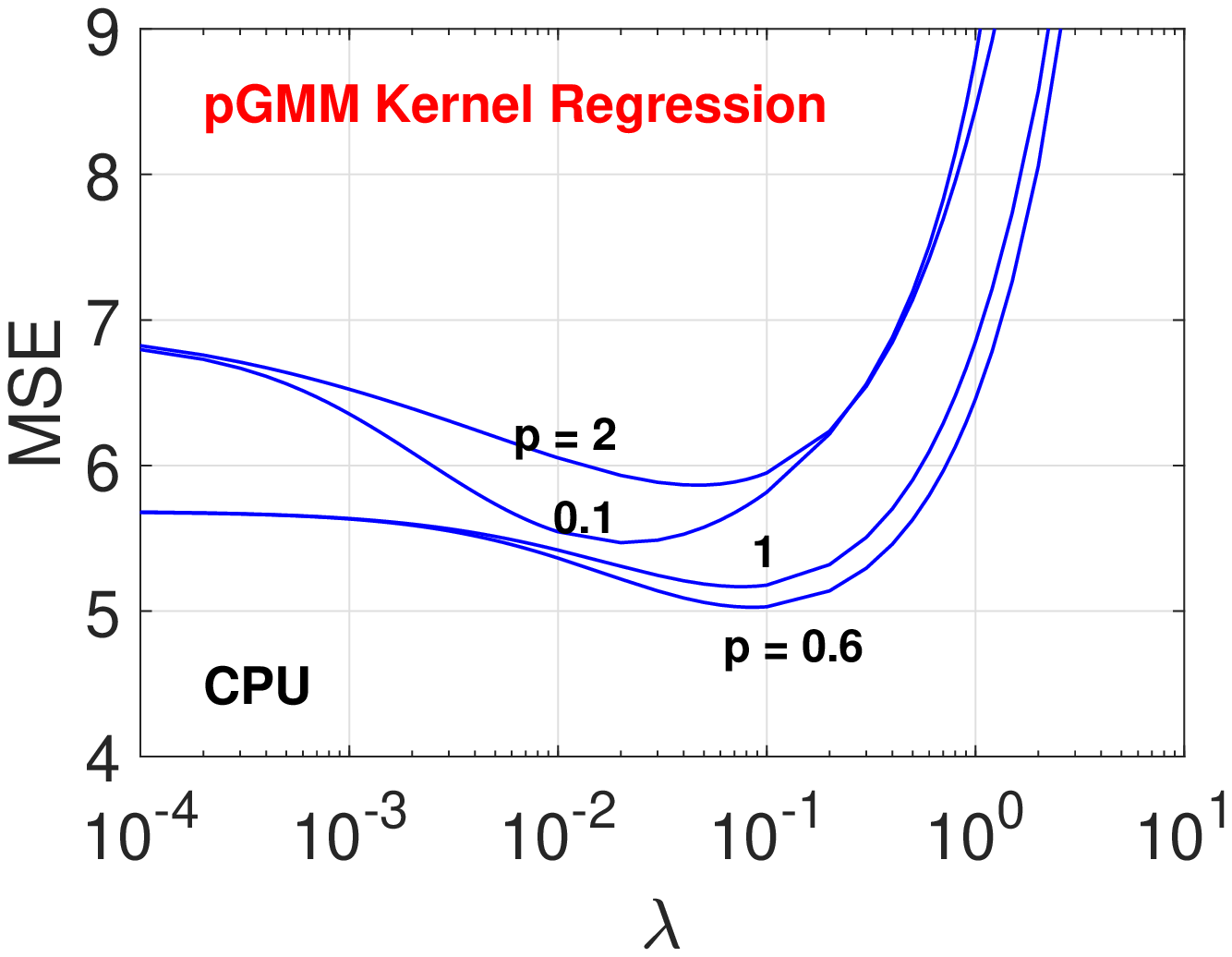}
}

\mbox{
    \includegraphics[width=2.7in]{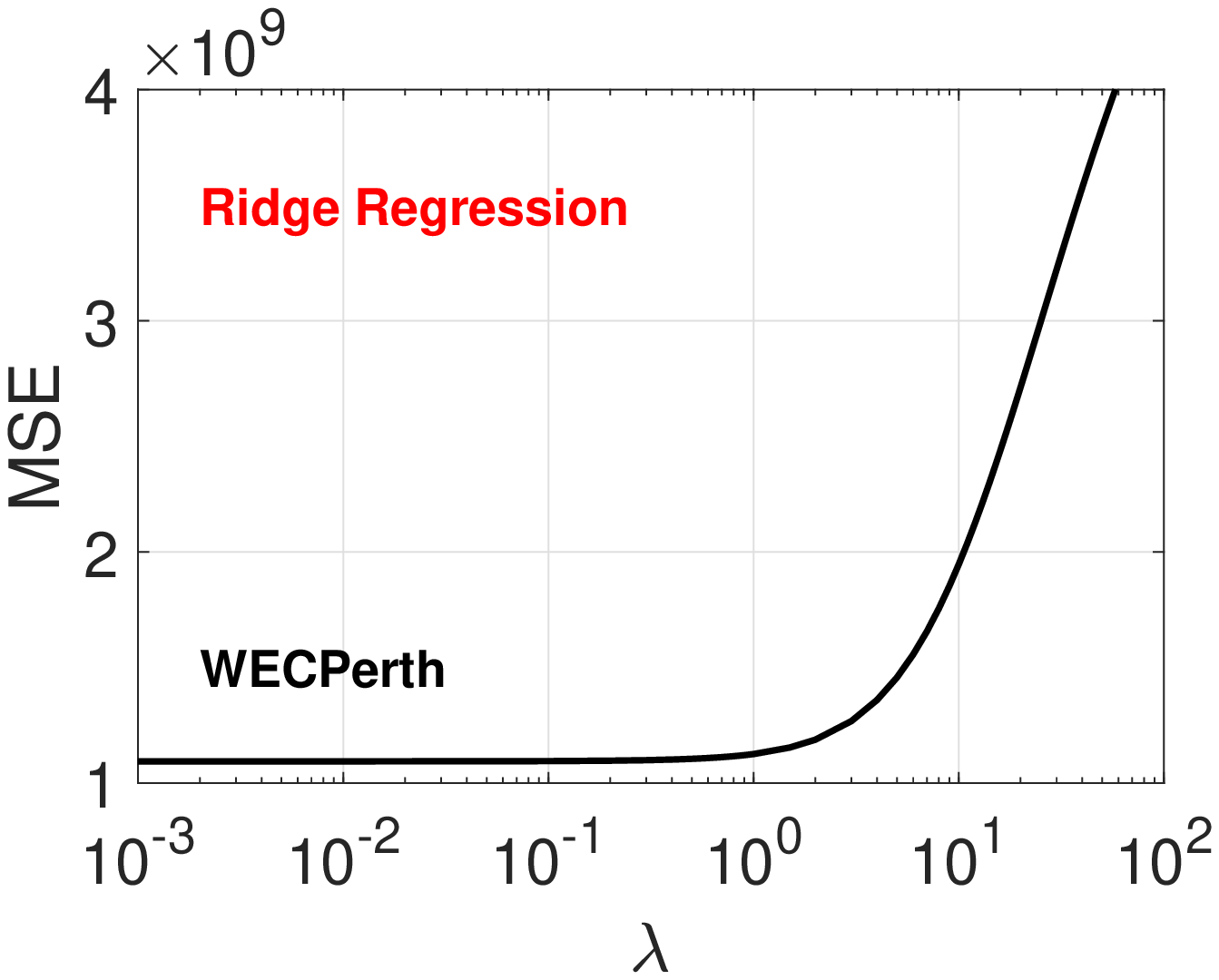}
    \includegraphics[width=2.7in]{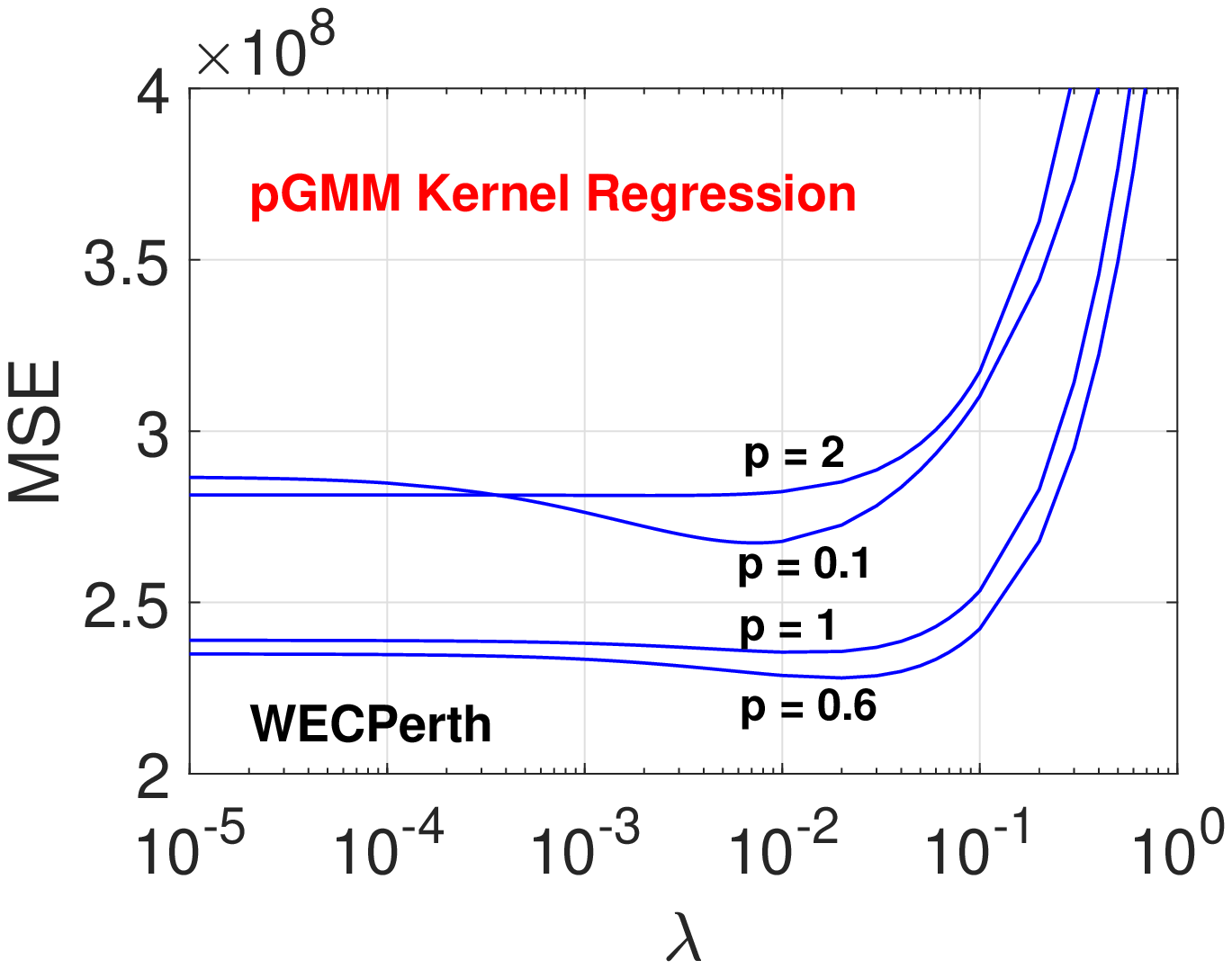}
}

\mbox{
    \includegraphics[width=2.7in]{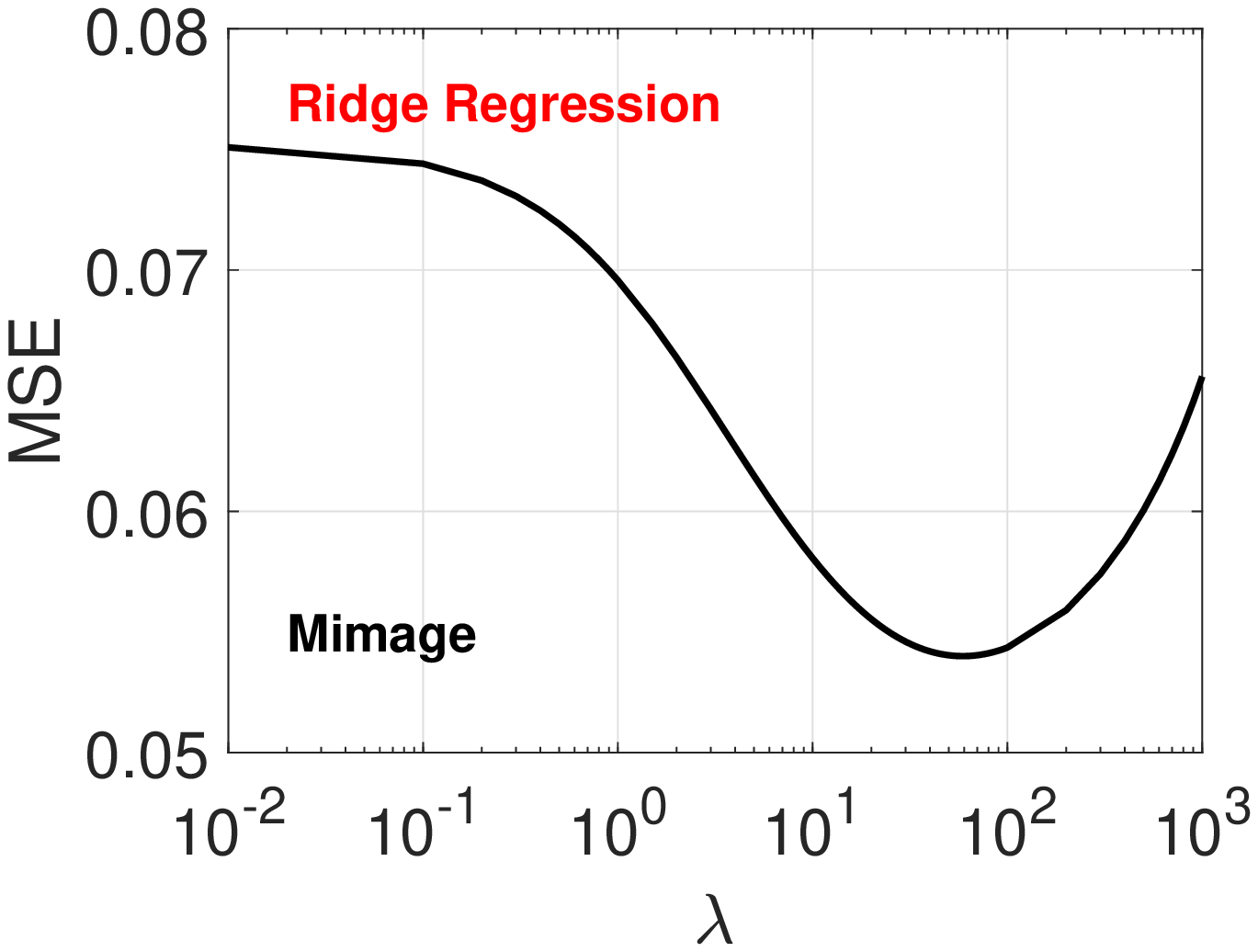}
    \includegraphics[width=2.7in]{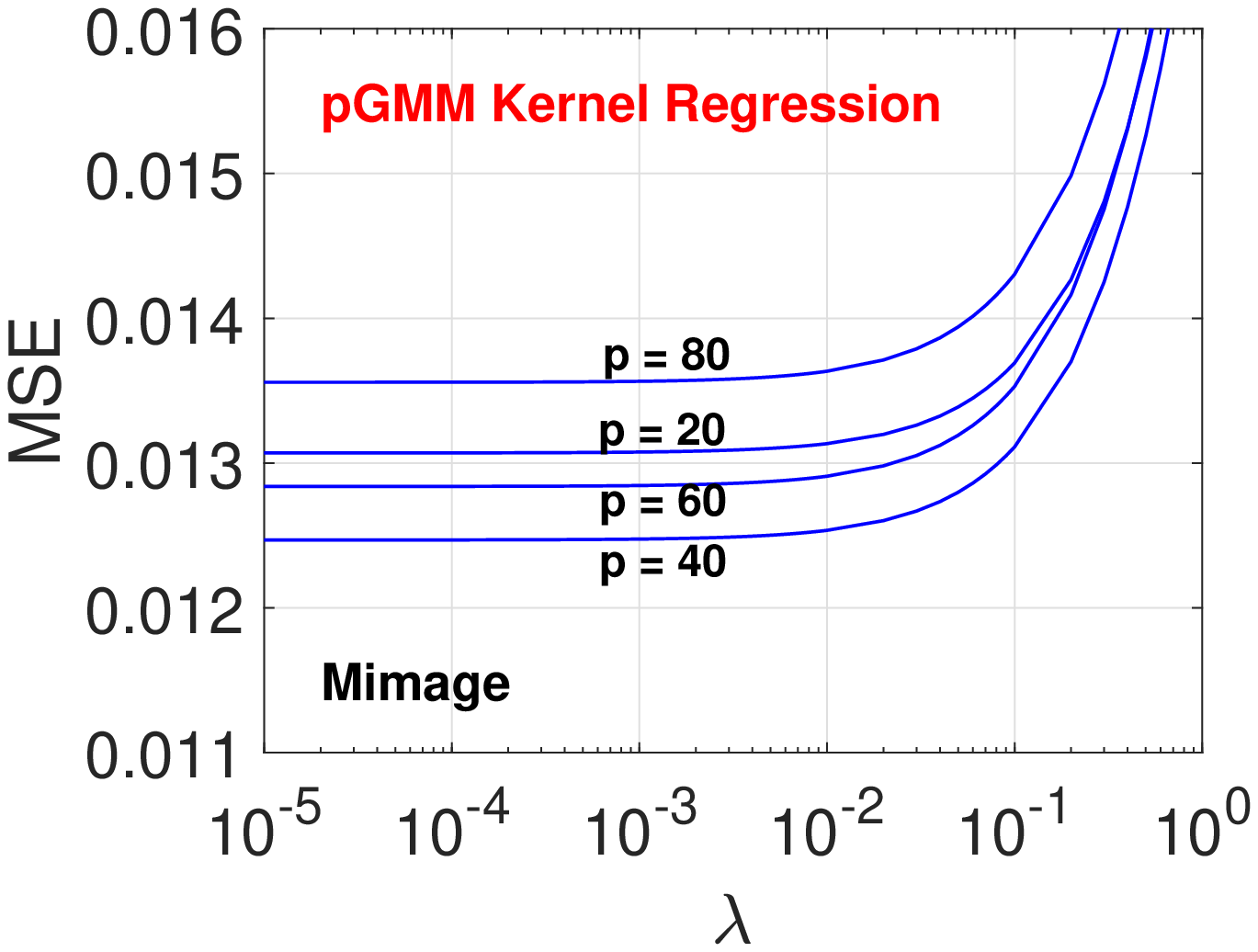}
}

\end{center}

    \caption{Test MSEs for selected datasets with respect to the ridge regression regularization $\lambda$, for ordinary ridge linear regression (LR, left panels) and pGMM kernel regression (right panels). }
    \label{fig:MSE_lambda}
\end{figure}

For ``LR'' (ordinary ridge linear regression), we report the best test (i.e., lowest) MSE over a wide range of the regularization coefficients $\lambda$. The left column of Figure~\ref{fig:MSE_lambda} plots, for selected datasets, the MSEs with respect to $\lambda$. For ``RBF'' (Gaussian kernel ridge regression), we report the best test MSEs over a wide range of regularization coefficients and the RBF kernel parameters. As expected,  the RBF kernel regression achieves substantially lower MSEs than LR. Note that (only) for LR we added the  ``intercept'' (i.e., an additional ``1'' in the data vector) which slightly improves LR.

Unlike the RBF kernel, the GMM kernel has no tuning parameter (i.e., the parameter $p=1$). Perhaps surprisingly, as reflected in Table~\ref{tab:data}, this tuning-free GMM kernel outperforms the best-tuned RBF kernel in most datasets, in some datasets quite substantially so. By adding a tuning parameter $p$, the pGMM kernel further improves the performance for essentially all datasets. The right column of Figure~\ref{fig:MSE_lambda} provides more details by plotting the MSEs of the pGMM kernel, for a wide range of regularization coefficients and selected $p$ values, for selected datasets.

\newpage\clearpage

Finally, the last column of Table~\ref{tab:data} reports the test MSEs of $L_2$ boost, which is implemented using the same codebase of ``Fast ABC-Boost''~\citep{li2022fast,li2022package}. Interestingly, we can see that, while $L_2$ boost often improves over the pGMM kernel, the improvements are usually quite small. This observation might be also surprising because one would expect complicated boosted tree models should substantially improve this simple one-parameter-only pGMM kernel. Of course, the prior studies in~\citet{li2017tunable,li2018several} have already demonstrated the same phenomenon, i.e., the tunable GMM kernels (including the pGMM kernel) perform quite close to boosted trees.

The next section will be devoted to $L_p$ boosting which often improves $L_2$ boost. To conclude this section, we should mention that the pGMM kernel might be further improved by introducing more tuning parameters as shown in~\citet{li2017tunable,li2018several}, or by combination with other types of kernels such as the strategy in the design of the ``CoRE Kernel''~\citep{li2014core} (which combines the cosine kernel with the (0/1)-min-max kernel).

\section{Regression by $L_p$ Boosting}

Boosting algorithms~\citep{schapire1990strength,freund1995boosting,freund1997decision,bartlett1998boosting,schapire1999improved,friedman2000additive,friedman2001greedy} are popular in machine learning research and applications. It is the standard practice to integrate boosting with trees~\citep{brieman1983classification} to produce accurate and robust prediction results. In this paper, we adopt the same codebase of ``Fast ABC-Boost'' (fast adaptive base class boost)~\citep{li2022fast,li2022package} to implement $L_p$ boost:
\begin{enumerate}
    \item The explicit gain formula for tree split using second-order information is re-derived for $L_p$ boosting regression. The same type of explicit second-order formula  as derived in~\citet{li2010robust} for classification has become the standard implementation in popular tree platforms.
    \item We use the same adaptive binning strategy developed in~\cite{li2007mcrank} to preprocess the features into discrete integer values.  This type of histogram-based implementation of trees is also common in popular tree platforms. As reported in~\cite{li2022package}, this simple (fixed-length) binning method is still surprisingly effective, compared with more sophisticated binning algorithms implemented in other tree platforms.
\end{enumerate}
To learn more about the ABC-Boost package, readers are also welcome to consult the following links (lecture notes, tutorials, and discussions):
\\\\~\url{http://statistics.rutgers.edu/home/pingli/STSCI6520/Lecture/ABC-LogitBoost.pdf} .\\
\url{http://www.stat.rutgers.edu/home/pingli/doc/PingLiTutorial.pdf} (pages 15--77).\\
\url{https://hunch.net/?p=1467} .

\subsection{$L_p$ Boosting  for Regression Using Second-Order Information}

We denote a training dataset by $\{y_i,\mathbf{x}_i\}_{i=1}^n$, where $n$ is the number of training samples, $\mathbf{x}_i$ is the $i$-th feature vector, and  $y_i$ is the $i$-th observation value. Our goal is to build a model $F(\mathbf{x})$ to minimize the $L_p$ loss  as follows:
\begin{align}\label{eqn:Lp}
L_p = \frac{1}{n}\sum_{i=1}^n L_{p,i} = \frac{1}{n}\sum_{i=1}^n |y_i - F_i|^p, \hspace{0.2in}\text{where } F_i = F(\mathbf{x}_i).
\end{align}
Following the ``additive model'' paradigm~\citep{friedman2000additive,friedman2001greedy},  we let $F$ be a function of $M$ terms:
\begin{align}\label{eqn_F_M}
F^{(M)}(\mathbf{x}) = \sum_{m=1}^M f_m(\mathbf{x}),
\end{align}
where  $f_m(\mathbf{x})$, the base learner, is  a regression tree and learned from the data in a stagewise greedy fashion. Following the idea from~\citet{friedman2000additive}, at each boosting iteration, we fit $f_m$ by weighted least squares, with  responses $\{z_i\}$ and weights $\{w_i\}$:
\begin{align}\label{eqn:zw}
&z_i = \frac{-L_{p,i}^\prime}{L_{p,i}^{\prime\prime}},\hspace{0.2in} w_i = L_{p,i}^{\prime\prime},\hspace{0.2in} \\\notag
&\text{where} \ \ L_{p,i}^\prime = \frac{\partial{L_{p,i}}}{\partial F_i}  \ \ \text{and } \  L_{p,i}^{\prime\prime} = \frac{\partial^2{L_{p,i}}}{\partial F_i^2}.
\end{align}
We adopt the derivation from~\cite{li2010robust} to obtain the corresponding \textbf{gain} formula  needed for deciding the split location in building regression trees using  responses $\{z_i\}$ and weights $\{w_i\}$.  Historically, boosting based on the weighted least square procedure  was believed to have numerical issues~\citep{friedman2000additive,friedman2008response},  and hence later~\citet{friedman2001greedy} proposed using only the first derivatives to fit the trees, i.e.,
\begin{align}\label{eqn:zw1}
z_i = -L_{p,i}^\prime, \hspace{0.2in} w_i=1.
\end{align}
Of course, it is now clear that, as shown in~\cite{li2010robust}, one can derive the explicit and numerically stable/robust formula for computing the gains using second-order information.

\vspace{0.1in}
Specifically, for the $L_p$ loss~\eqref{eqn:Lp}, we can write down its first derive, for $p\geq 1$:
\begin{align}\label{eqn:dL}
\frac{\partial L_{p,i}}{\partial F_i}  =   -p|y_i-F_i|^{p-1}\text{sign}\left(y_i-F_i\right),
\end{align}
and the second derivative, for $p\geq 2$:
\begin{align}\label{eqn:d2L}
\frac{\partial^2 L_{p,i}}{\partial F_i^2}  =   p(p-1)|y_i-F_i|^{p-2}.
\end{align}
Note that $\frac{\partial^2 L_{p,i}}{\partial F_i^2} =2$ when $p=2$.

\subsection{Tree-Splitting Criterion Using Second-Order Information}\label{sec_split}

Here we repeat derivations in~\cite{li2010robust}. Consider a tree node with $N$ data points and consider one particular feature.  We have the weights $w_i$ and the response values $z_i$, $i=1$ to $N$. The data points are already sorted according to the feature values. The tree-splitting procedure is to find the index $s$, $1\leq s<N$, such that the weighted  square error (SE) is reduced the most if split at $s$.

\newpage

That is, we seek the $s$ to maximize
\begin{align}\notag
Gain(s) = &SE_{total} - (SE_{left} + SE_{right})\\\notag
=&\sum_{i=1}^N (z_i - \bar{z})^2w_i - \left[
\sum_{i=1}^s (z_i - \bar{z}_L)^2w_i + \sum_{i=s+1}^N (z_i - \bar{z}_R)^2w_i\right],
\end{align}
where $\bar{z} = \frac{\sum_{i=1}^N z_iw_i}{\sum_{i=1}^N w_i}$,
$\bar{z}_{left} = \frac{\sum_{i=1}^s z_iw_i}{\sum_{i=1}^s w_i}$,
$\bar{z}_{right} = \frac{\sum_{i=s+1}^N z_iw_i}{\sum_{i=s+1}^{N} w_i}$.

With some algebra, one can obtain
\begin{align}\notag
\sum_{i=1}^N (z_i - \bar{z})^2w_i =& \sum_{i=1}^N z_i^2w_i + \bar{z}^2w_i-2z_i\bar{z}w_i\\\notag
=&\sum_{i=1}^N z_i^2w_i + \left(\frac{\sum_{i=1}^N z_iw_i}{\sum_{i=1}^N w_i}\right)^2\sum_{i=1}^N w_i - 2\frac{\sum_{i=1}^N z_iw_i}{\sum_{i=1}^N w_i}\sum_{i=1}^N z_iw_i\\\notag
=&\sum_{i=1}^N z_i^2w_i - \frac{\left(\sum_{i=1}^N z_iw_i\right)^2}{\sum_{i=1}^N w_i}\\\notag
\sum_{i=1}^s (z_i - \bar{z}_{left})^2w_i
=&\sum_{i=1}^s z_i^2w_i - \frac{\left(\sum_{i=1}^s z_iw_i\right)^2}{\sum_{i=1}^t w_i}\\\notag
\sum_{i=s+1}^N (z_i - \bar{z}_{right})^2w_i
=&\sum_{i=s+1}^N z_i^2w_i - \frac{\left(\sum_{i=s+1}^N z_iw_i\right)^2}{\sum_{i=s+1}^N w_i}
\end{align}
Therefore, after simplification, we obtain
\begin{align}\notag
Gain(s) =& \frac{\left[\sum_{i=1}^s z_iw_i\right]^2}{\sum_{i=1}^s w_i}+\frac{\left[\sum_{i=s+1}^N z_iw_i\right]^2}{\sum_{i=s+1}^{N} w_i}- \frac{\left[\sum_{i=1}^N z_iw_i\right]^2}{\sum_{i=1}^N w_i}
\end{align}
Plugging in  $z_i = -L_i^{\prime}/L_i^{\prime\prime}$, $w_i = L_i^{\prime\prime}$  yields,
\begin{align}\label{eqn:gain}
Gain(s) =&  \frac{\left[\sum_{i=1}^s L_i^\prime \right]^2}{\sum_{i=1}^s L_i^{\prime\prime}}+\frac{\left[\sum_{i=s+1}^N L_i^{\prime}\right]^2}{\sum_{i=s+1}^{N} L_i^{\prime\prime}}- \frac{\left[\sum_{i=1}^N  L_i^\prime \right]^2}{\sum_{i=1}^N L_i^{\prime\prime}}.
\end{align}
This procedure is numerically robust/stable because we never need to directly compute the response values $z_i =- L_i^\prime/L_i^{\prime\prime}$, which can (and should) approach infinity easily. Because the original LogitBoost~\citep{friedman2000additive} directly used the individual response values  $z_i =- L_i^\prime/L_i^{\prime\prime}$, the procedure was believed to have numerical issues, which was one motivation for~\cite{friedman2001greedy} to use only the first derivatives to build tress i.e., $z_i = L_i^\prime$ $w_i =1$.  Thus the gain formula becomes:
\begin{align}\label{eqn:ugain}
UGain(s) =&  \frac{1}{s}\left[\sum_{i=1}^s L_i^\prime \right]^2+
\frac{1}{N-s}\left[\sum_{i=s+1}^N  L_i^\prime \right]^2-\frac{1}{N}
\left[\sum_{i=1}^N  L_i^\prime \right]^2.
\end{align}

Note that in the above derivations, we simply use $L_i$, $L_i^\prime$, and $L_i^{\prime\prime}$ by dropping the subscript $p$ (e.g., $L_{p,i}$) because the formulas are really applicable to general loss functions, not just the $L_p$ loss.

\newpage

\subsection{$L_p$ Boosting Algorithm }

Algorithm~\ref{alg:LpBoost} describes Robust LogitBoost using the tree split gain formula~\eqref{eqn:gain} (for $p\geq 2$) or the tree split gain formula~\eqref{eqn:ugain} (for $1\leq p<2$). Note that after trees are constructed, the values of the terminal nodes are computed by
\begin{align}\notag
\frac{\sum_{node} z_{i,k} w_{i,k}}{\sum_{node} w_{i,k}} =
\frac{\sum_{node}-L_i^\prime}{\sum_{node} L_i^{\prime\prime}},
\end{align}
which explains Line 5 of Algorithm~\ref{alg:LpBoost}.  When $1\leq p<2$, because the second derivative $L^{\prime\prime}_i$ does not exist, we follow~\citet{friedman2001greedy} by only using the first derivatives to build the trees using the tree split gain formula~\eqref{eqn:ugain}, and we update the note values by
\begin{align}\notag
\frac{\sum_{node} z_{i,k} w_{i,k}}{\sum_{node} w_{i,k}} =
\frac{\sum_{node}-L_i^\prime}{p \times \#|node|},
\end{align}
which explains Line 9 of Algorithm~\ref{alg:LpBoost}. Note that we use $p \times \#|node|$ instead of just $\times \#|node|$ to avoid potential ambiguity at $p=2$.

\vspace{0.1in}

In retrospect, deriving Eq.~\eqref{eqn:gain}, the explicit  tree split gain formula using  second-order information, is straightforward. Nevertheless, until the formula was first presented by~\cite{li2010robust}, the original LogitBoost was believed to have numerical issues and the more robust MART algorithm was a  popular alternative~\citep{friedman2000additive,friedman2008response}. As shown in~\citet{li2010robust}, in many datasets, because MART used only the first derivative information for tree split, MART typically did not achieve as good accuracy as Robust LogitBoost. Nowadays, the split gain formula in Eq.~\eqref{eqn:gain} is the standard implementation in popular tree platforms.  On the other hand, we should emphasize that it is not guaranteed that boosting using second derivatives would always improve over boosting using only first derivatives. It might be good to maintain the implementations of both approaches for practitioners to choose from~\citep{li2022package}.

{\begin{algorithm}[t]
$F_{i} = 0$, $i = 1$ to $n$ \\
For $m=1$ to $M$ Do\\
\hspace{0.15in}  If $p\geq 2$ Do\\
\hspace{0.3in}  $\left\{R_{j,m}\right\}_{j=1}^J = J$-terminal node weighted regression tree from
 $\{z_i = -L_i^\prime/L_i^{\prime\prime}, \ \ w_i = L_i^{\prime\prime}, \ \  \mathbf{x}_{i}\}_{i=1}^n$, using the tree split gain formula Eq.~\eqref{eqn:gain}.\\
 \hspace{0.3in}  $\beta_{j,m} = \frac{ \sum_{\mathbf{x}_i \in
  R_{j,m}} -L_i^\prime}{ \sum_{\mathbf{x}_i\in
  R_{j,m}} L_i^{\prime\prime} }$ \\
 \hspace{0.15in} End\vspace{0.1in}\\
\hspace{0.15in}  If $1\leq p< 2$ Do \\
\hspace{0.3in}  $\left\{R_{j,m}\right\}_{j=1}^J = J$-terminal node regression tree from
 $\{z_i = -L_i^\prime, \ \  \mathbf{x}_{i}\}_{i=1}^n$, using the tree split gain formula Eq.~\eqref{eqn:ugain}.\\
 \hspace{0.3in}  $\beta_{j,m} = \frac{ \sum_{\mathbf{x}_i \in
  R_{j,m}} -L_i^\prime}{
  p\times \#|R_{j,m}|}$ \\
 \hspace{0.15in} End\\
\hspace{0.15in}  $F_{i} = F_{i} +
\nu\sum_{j=1}^J\beta_{j,m}1_{\mathbf{x}_i\in R_{j,m}}$ \\
End
\caption{$L_p$ boosting. $L^\prime = -p|y_i-F_i|^{p-1}\text{sign}\left(y_i-F_i\right)$,  $L^{\prime\prime} = p(p-1)|y_i-F_i|^{p-2}$.}
\label{alg:LpBoost}
\end{algorithm}}

\newpage\clearpage

\subsection{Experiments}

\begin{figure}[h]
\begin{center}

\mbox{
    \includegraphics[width=2.2in]{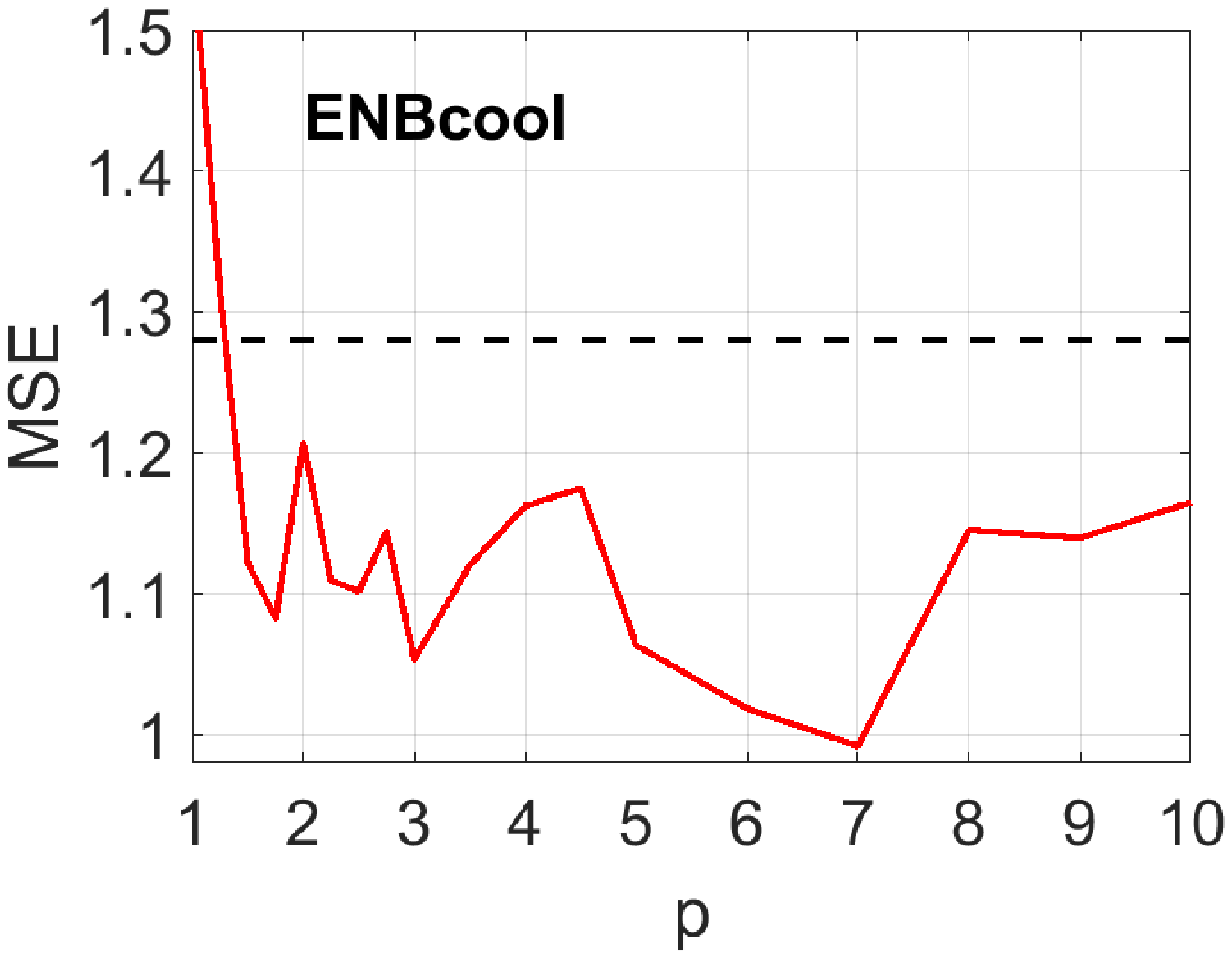}
    \includegraphics[width=2.2in]{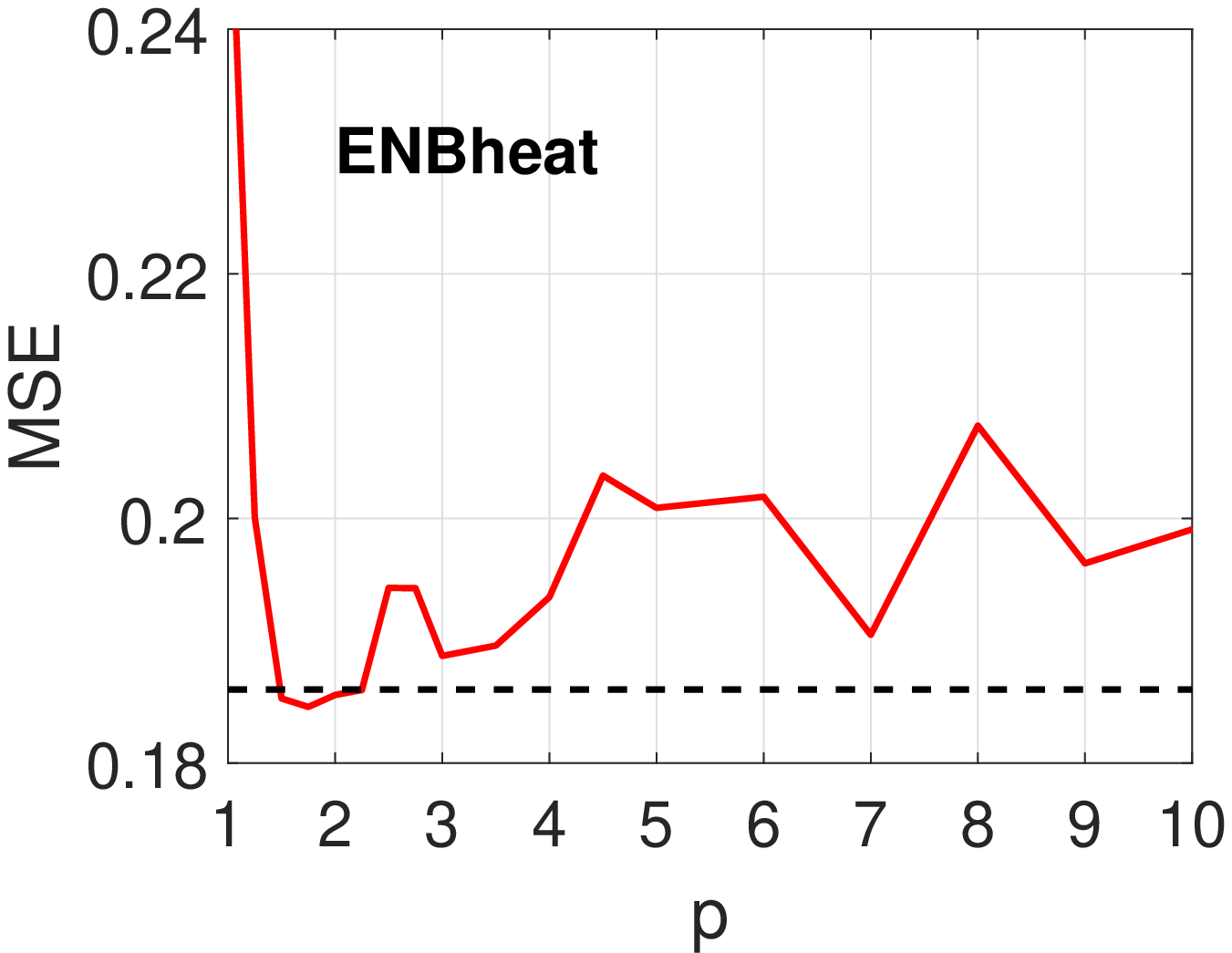}
    \includegraphics[width=2.2in]{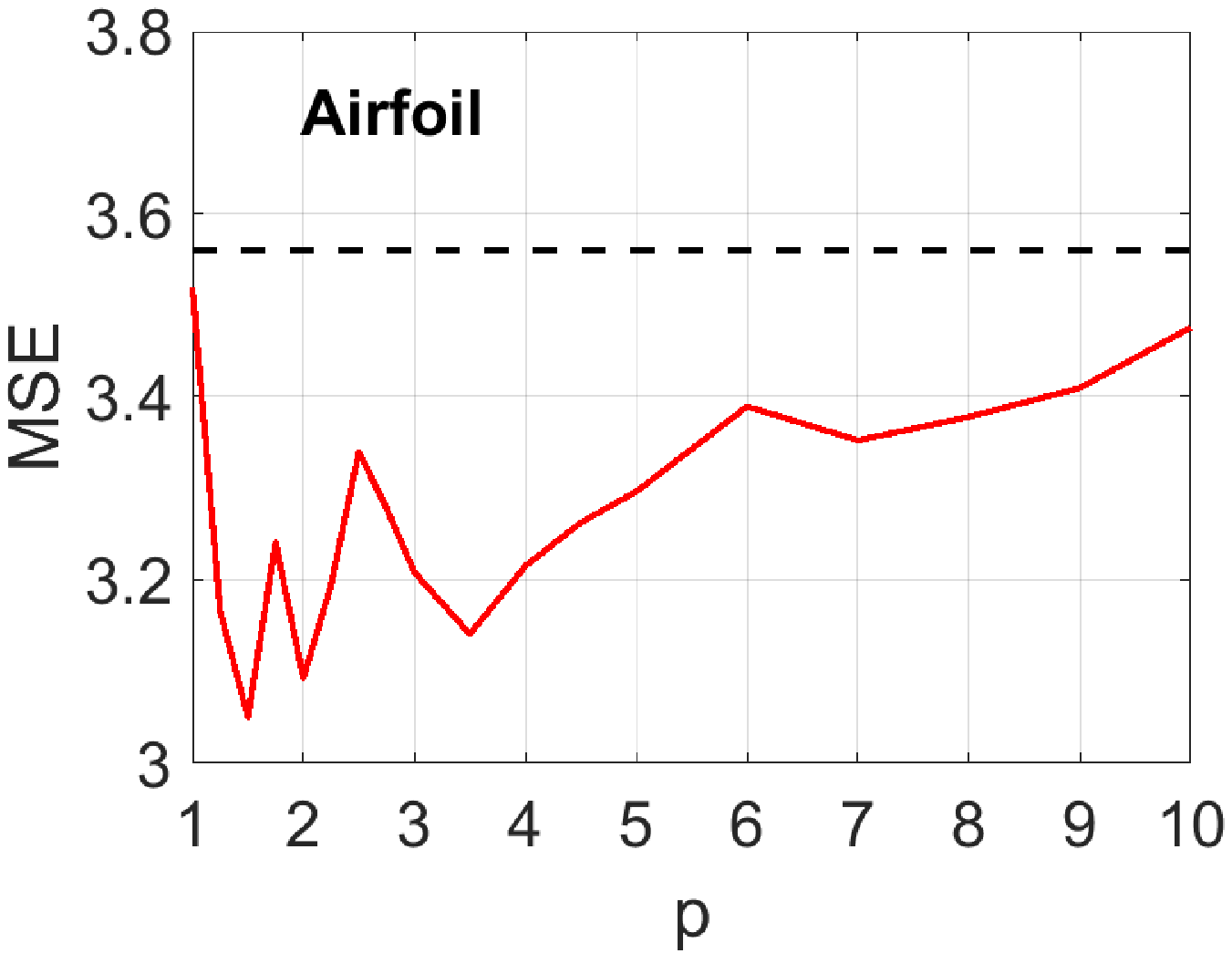}
}

\mbox{
    \includegraphics[width=2.2in]{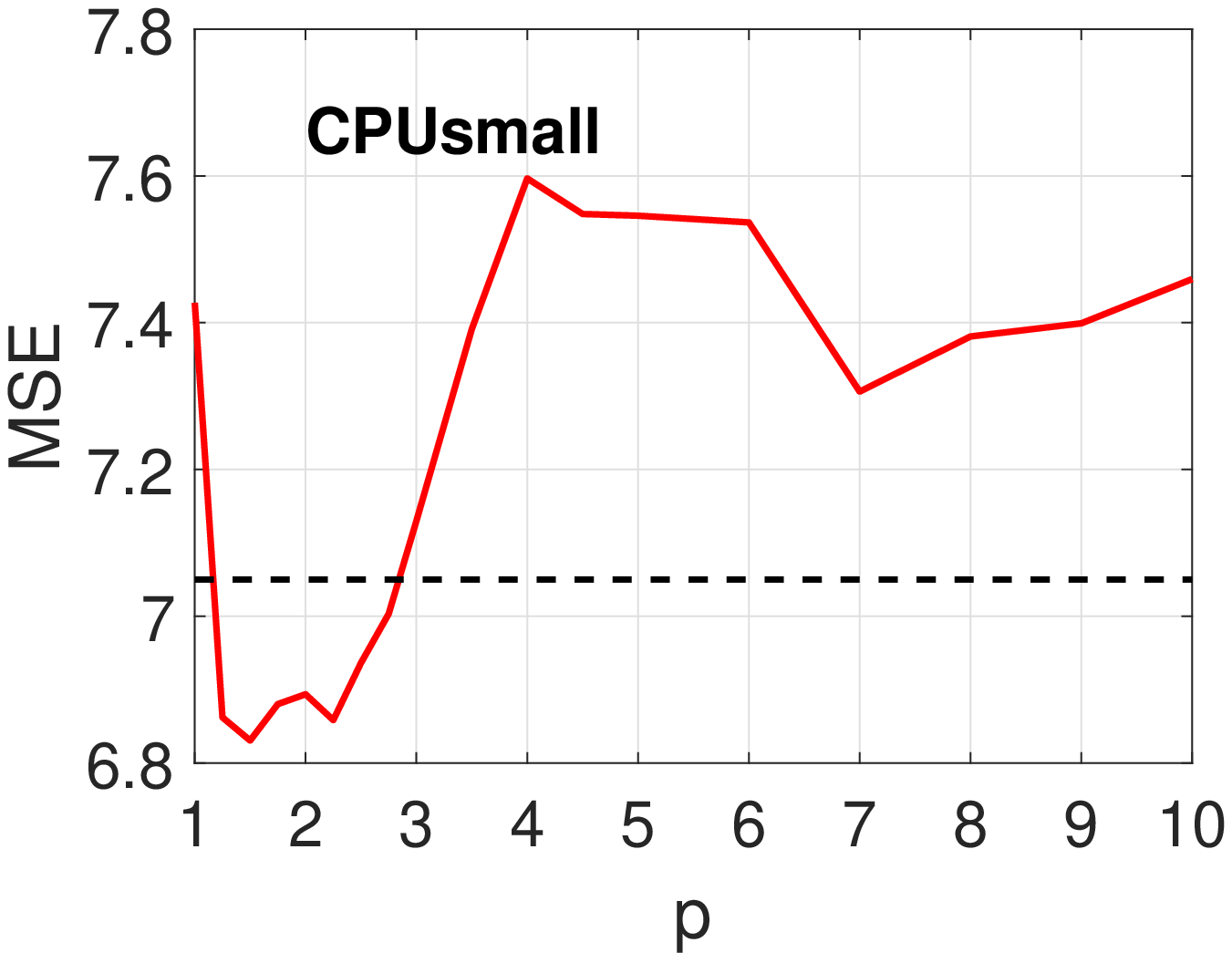}
    \includegraphics[width=2.2in]{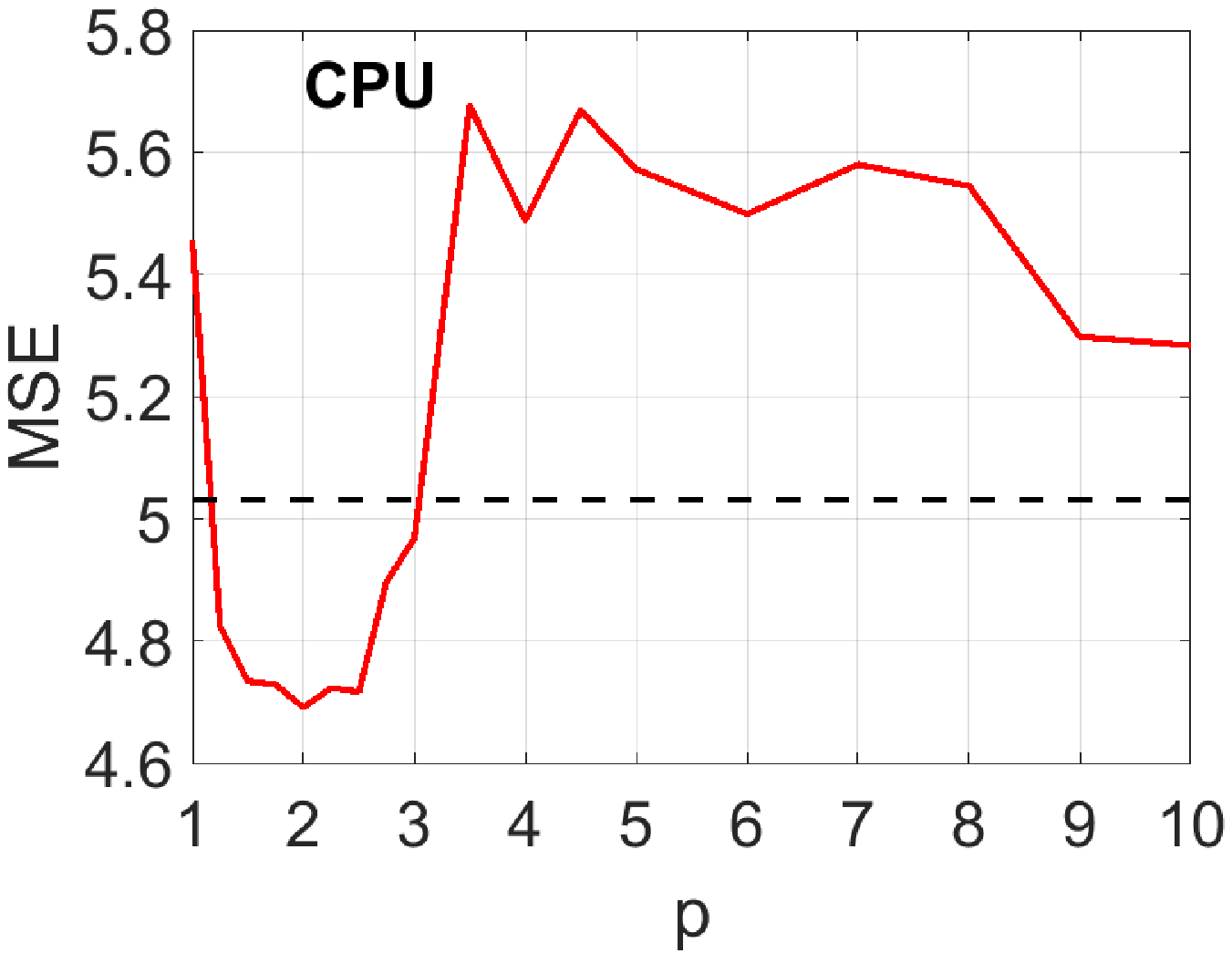}
    \includegraphics[width=2.2in]{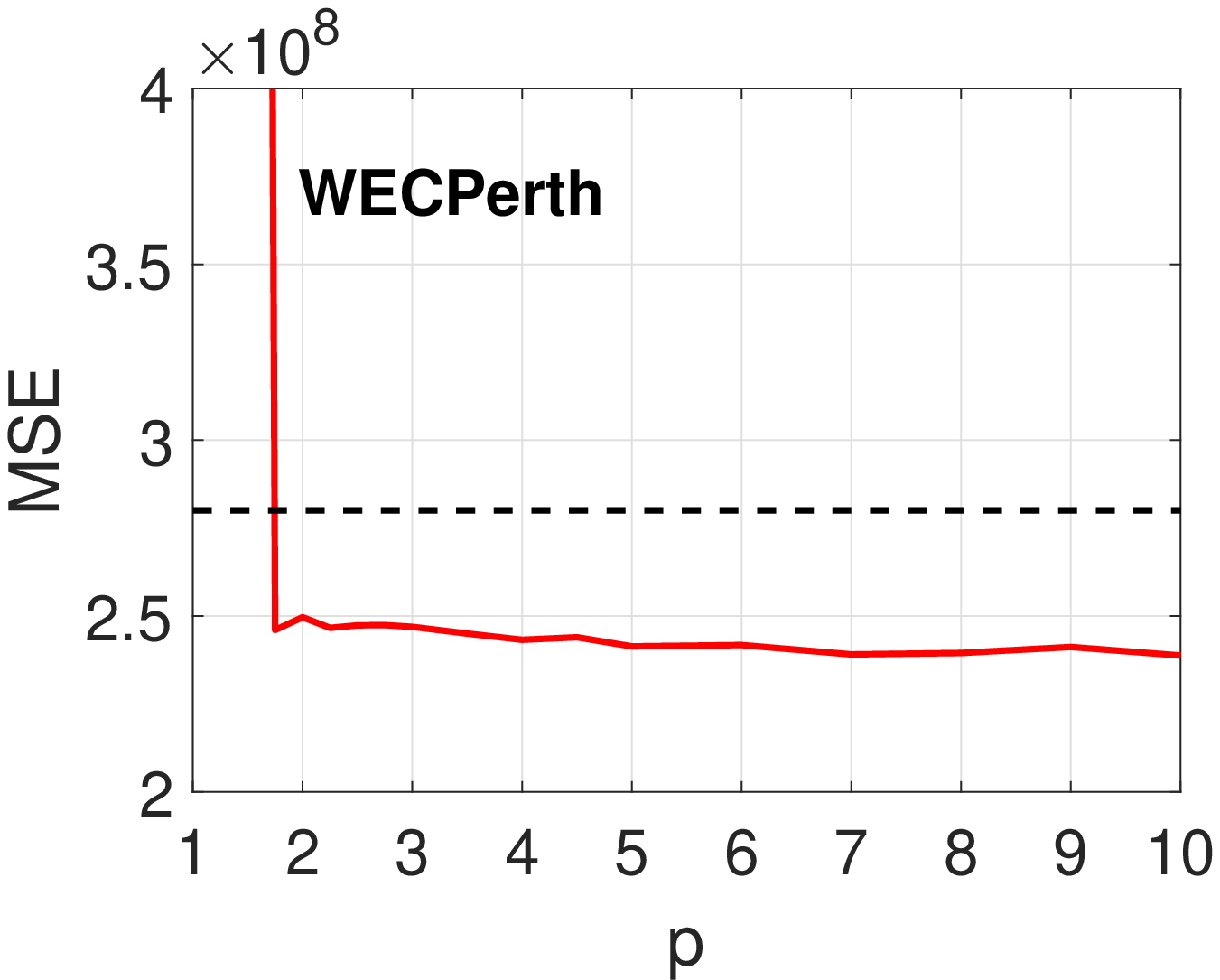}
}

\mbox{
    \includegraphics[width=2.2in]{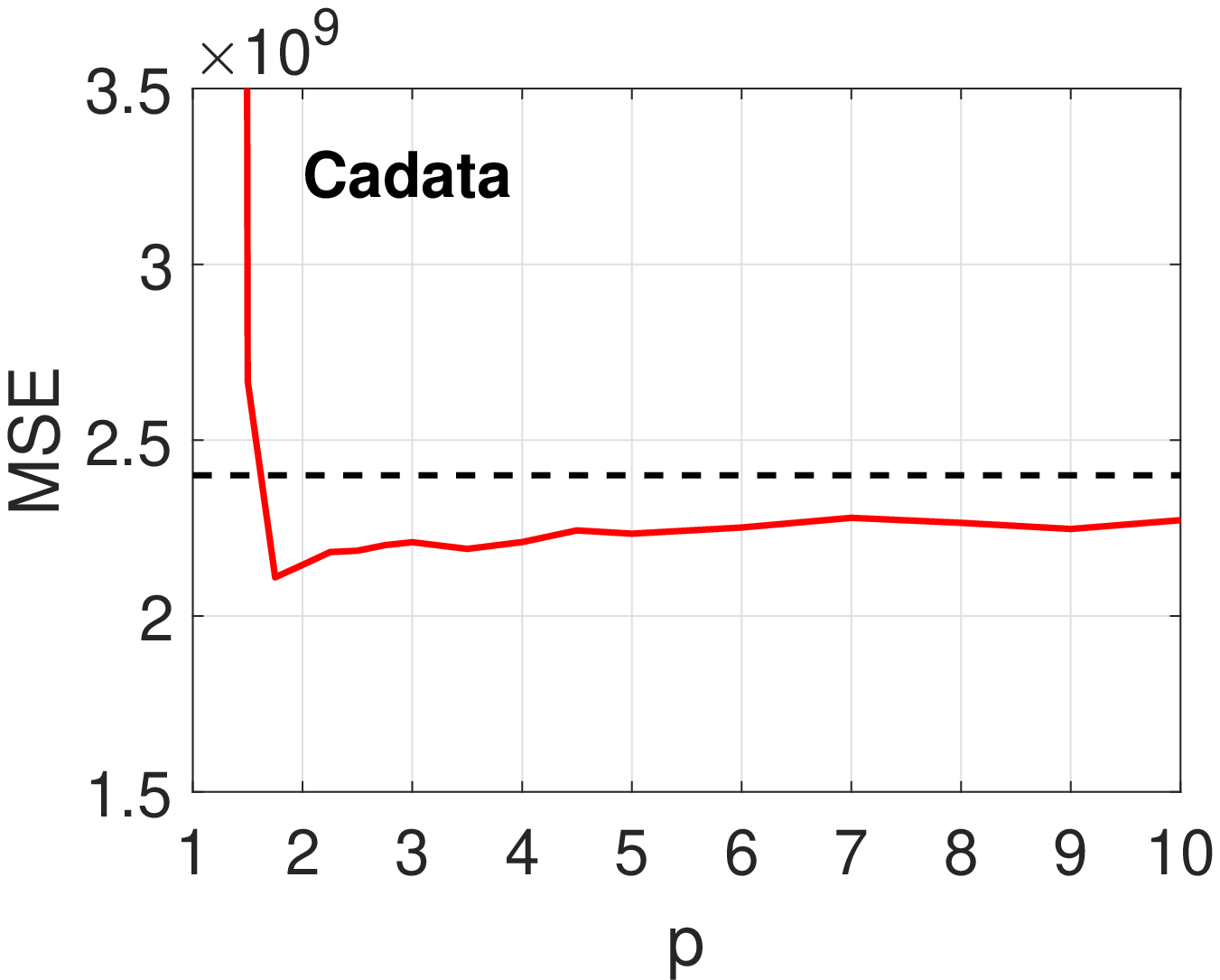}
    \includegraphics[width=2.2in]{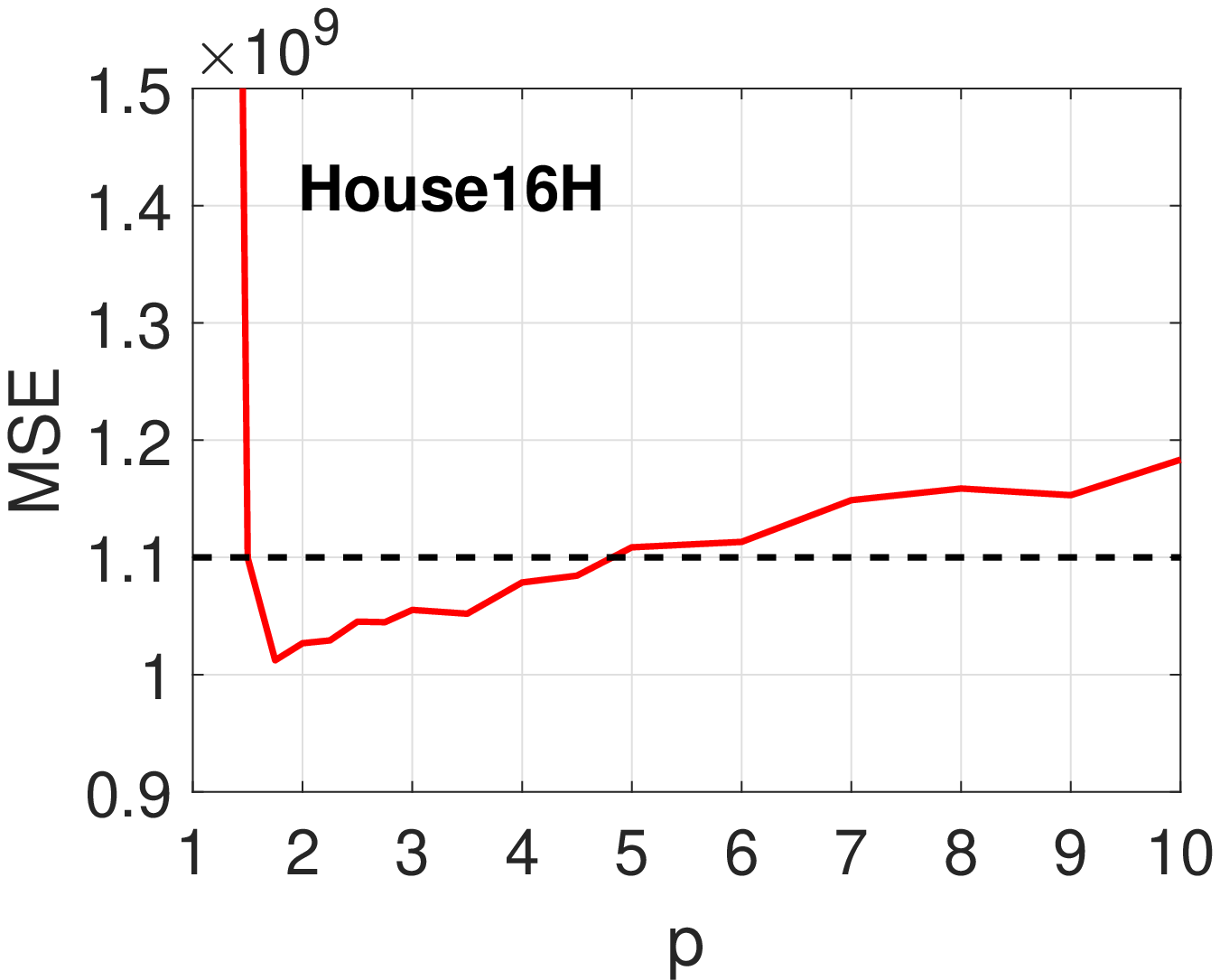}
    \includegraphics[width=2.2in]{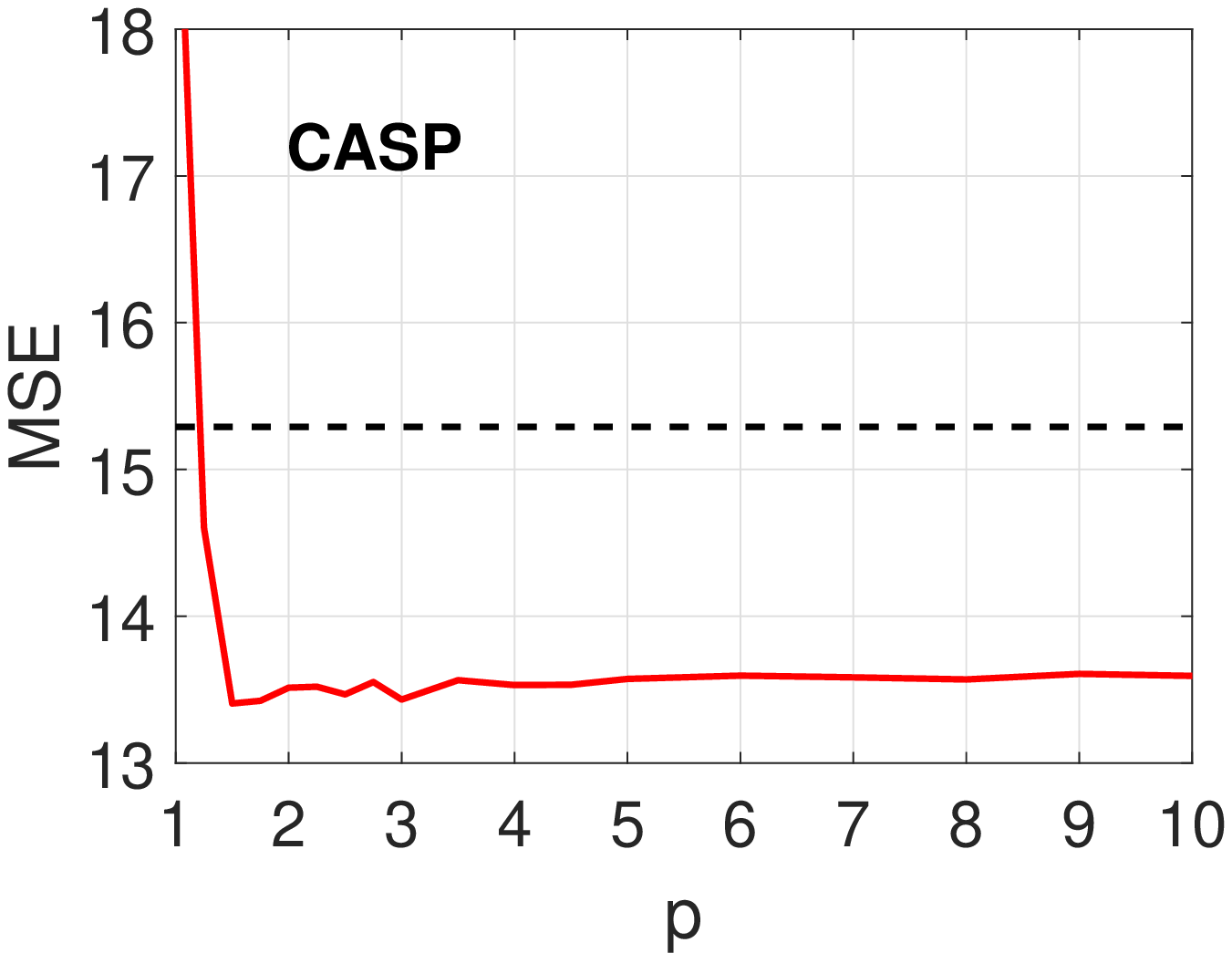}
}

\mbox{
    \includegraphics[width=2.2in]{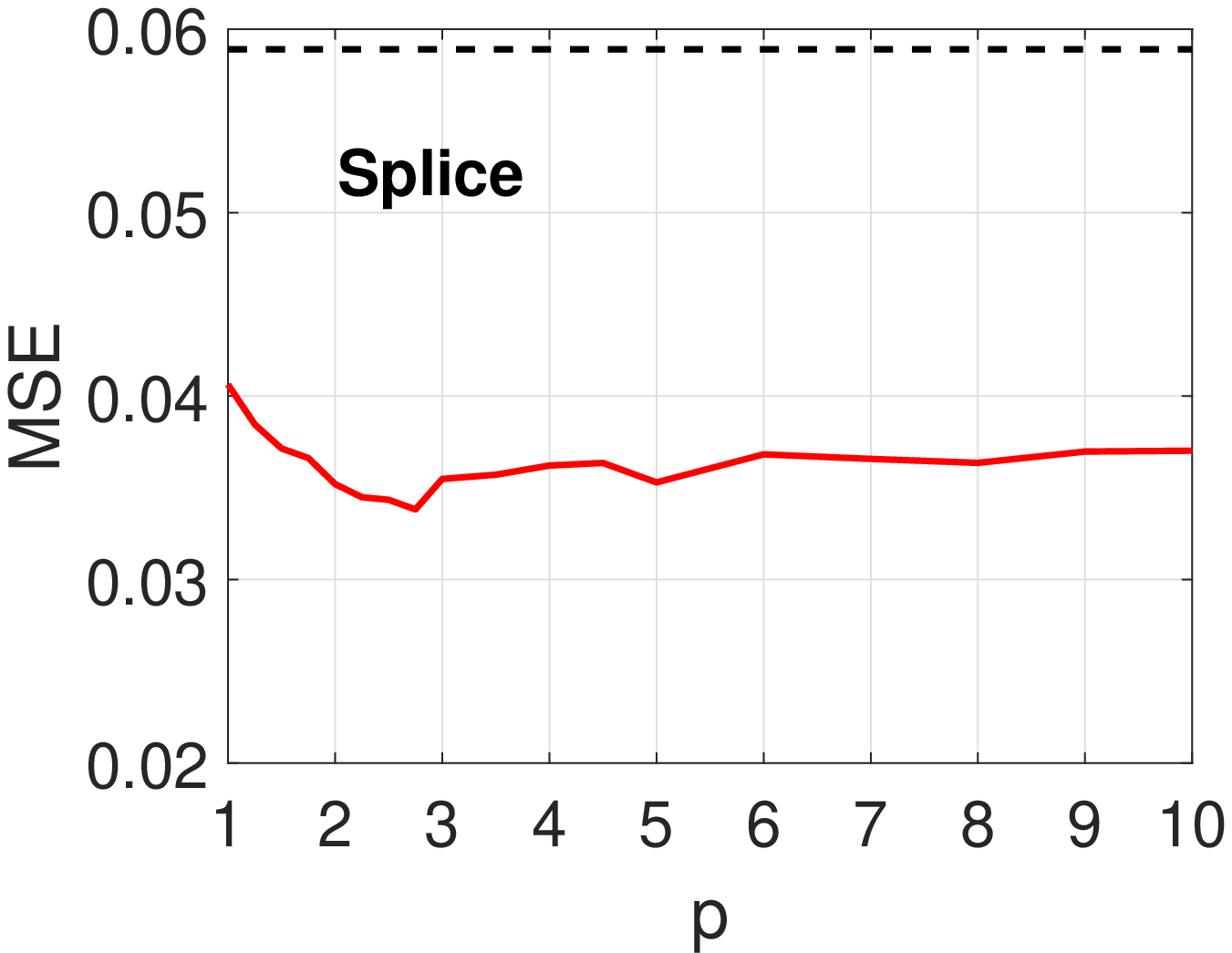}
    \includegraphics[width=2.2in]{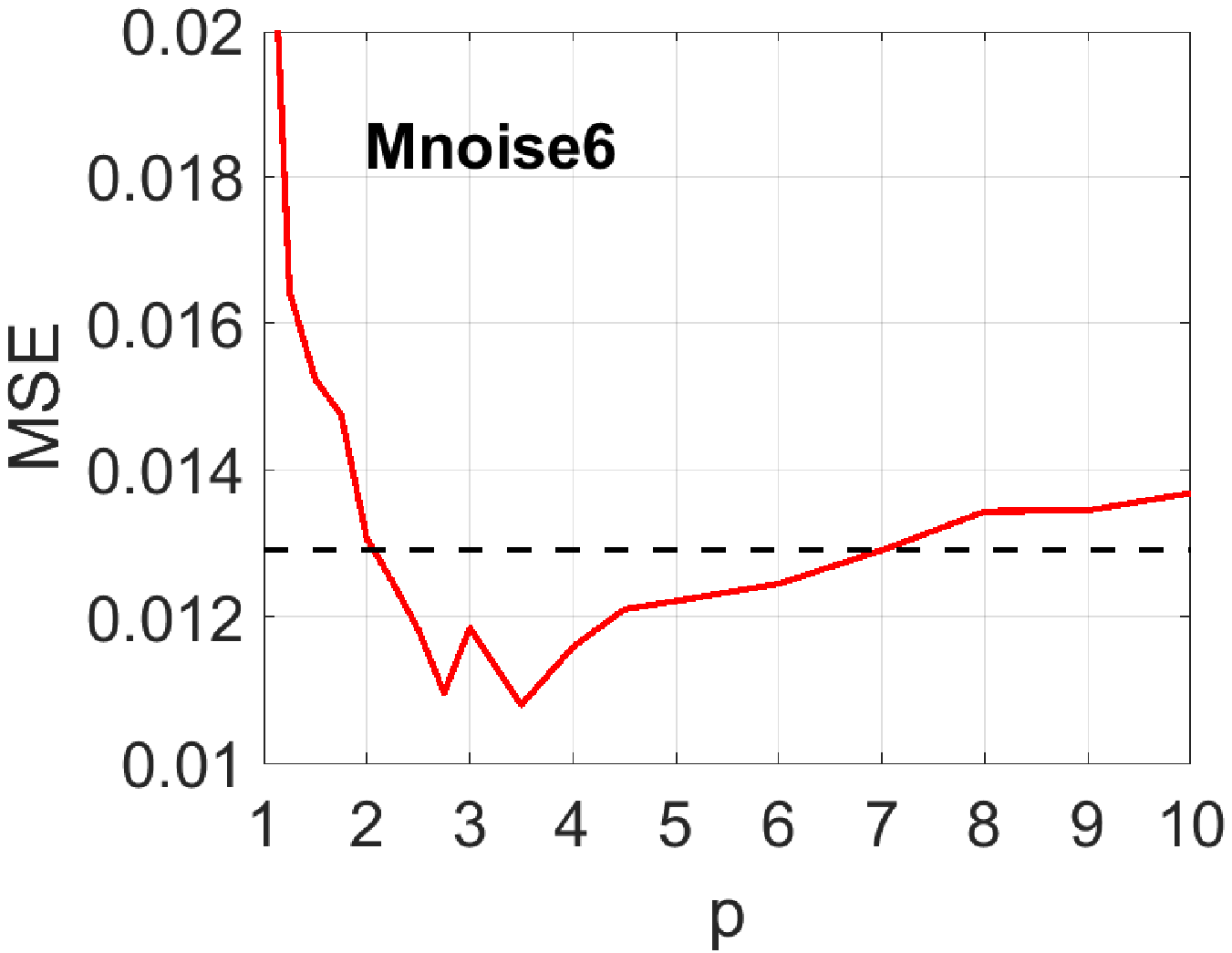}
    \includegraphics[width=2.2in]{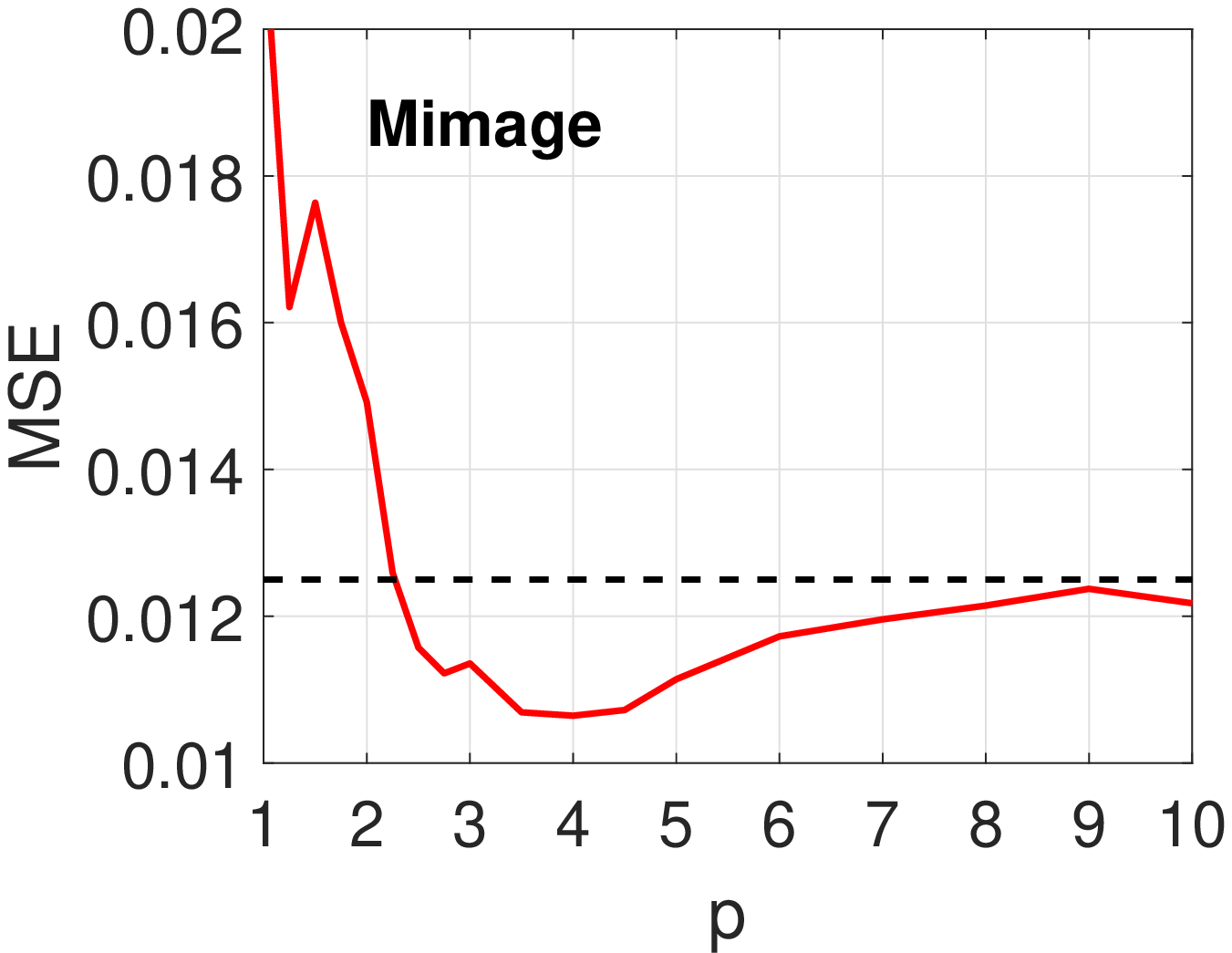}
}

\end{center}

    \caption{Test MSEs with respect to $p$, using $L_p$ boosting. The dashed horizontal lines represent the test MSEs of the pGMM kernel. $L_p$ boosting outperforms the pGMM kernel but differences are often not too large. Interestingly, for many datasets, $p=2$ is not the optimal choice in $L_p$ boosting.}
    \label{fig:LpMSE}
\end{figure}

We have implemented $L_p$ regression using the same codebase of ``Fast ABC-Boost''~\citep{li2022fast,li2022package} and we test it on the same datasets as listed in Table~\ref{tab:data}. How to evaluate $L_p$ regression and compare it with other algorithms is an interesting question. While the optimization procedure of $L_p$ regression is to minimize the $L_p$ loss
\begin{align}\notag
L_p =  \frac{1}{n}\sum_{i=1}^n |y_i - F_i|^p,
\end{align}
we decide to report the $L_2$ loss: $L_2 =  \frac{1}{n}\sum_{i=1}^n |y_i - F_i|^2$ in Figure~\ref{fig:LpMSE}, regardless of which $p$ we use in the $L_p$ regression. Note that in Figure~\ref{fig:LpMSE}, we also call the $L_2$ loss as the MSE in the y-axis. In the open-source code~\citep{li2022package}, we output the $L_2$ loss as well as the $L_p$ loss.

\vspace{0.1in}

We experiment with parameters $J\in\{6, 10, 20\}$, $\nu\in\{0.06, 0.1, 0.2\}$, and  $M=10000$ iterations. We adopt a conservative early stopping criterion and exit after the $L_p$ loss is lower than
\begin{align}\label{eqn:stop}
\epsilon^{p/2} \times \frac{1}{n}\sum_{i=1}^n{|y_i|^p},
\end{align}
where $\epsilon = 10^{-5}$ by default which can be adjusted by users.

Figure~\ref{fig:LpMSE} reports the best test MSEs overall $J$, $\nu$, and iterations, for each fixed $p$ value. We can see that, for some datasets,   $L_p$ boosting with $p=2$  indeed achieves (close to) the best (lowest) $L_2$ loss. However, for many other datasets, $L_p$ boost with $p>2$ achieves lower $L_2$ loss than $L_2$  boost. These results indicate that practitioners might be able to  effectively treat $p$ as a tuning parameter in order to achieve better $L_2$ loss.

\vspace{0.1in}

\begin{figure}[h]
\begin{center}

\mbox{
    \includegraphics[width=2.2in]{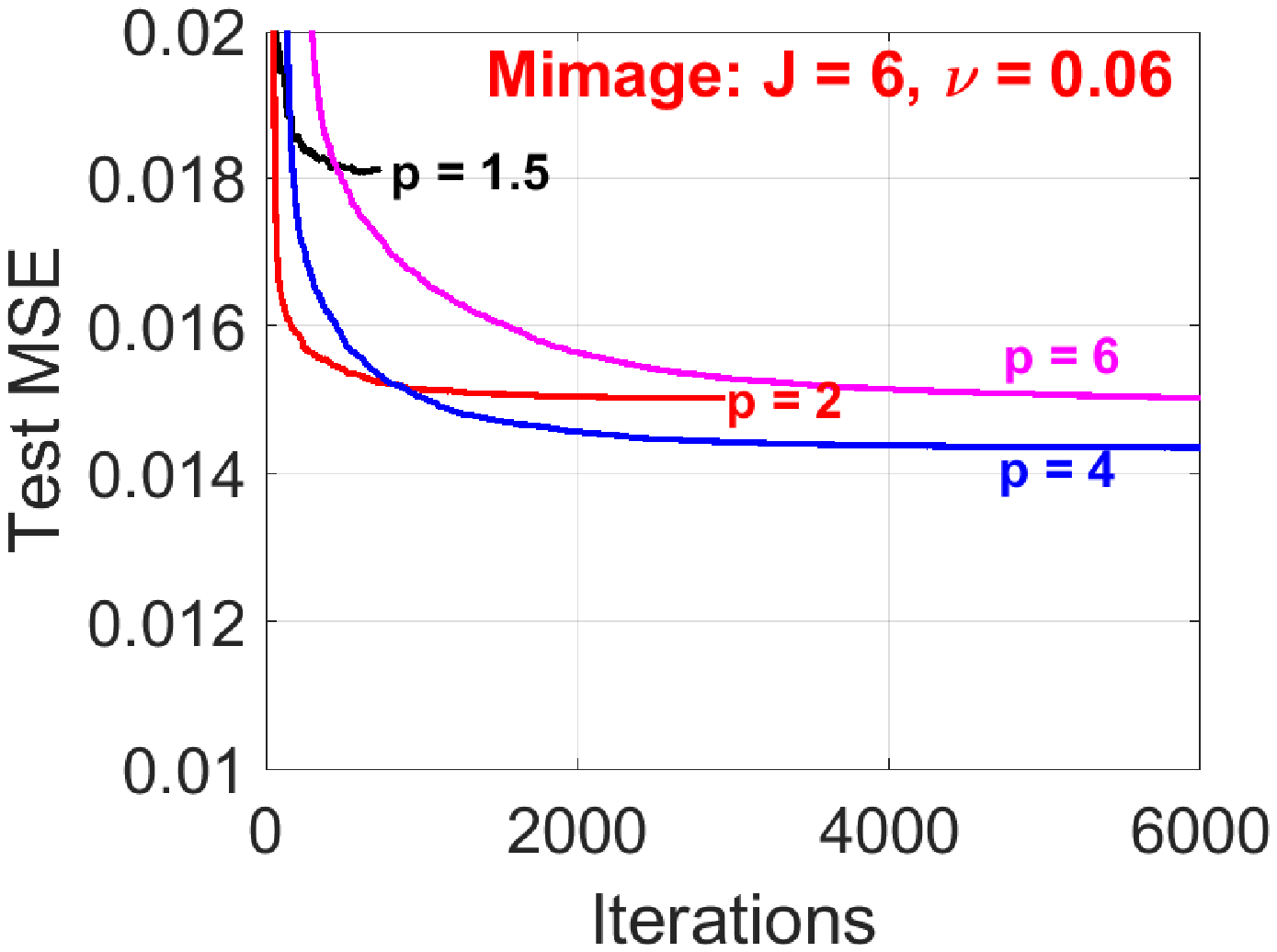}
    \includegraphics[width=2.2in]{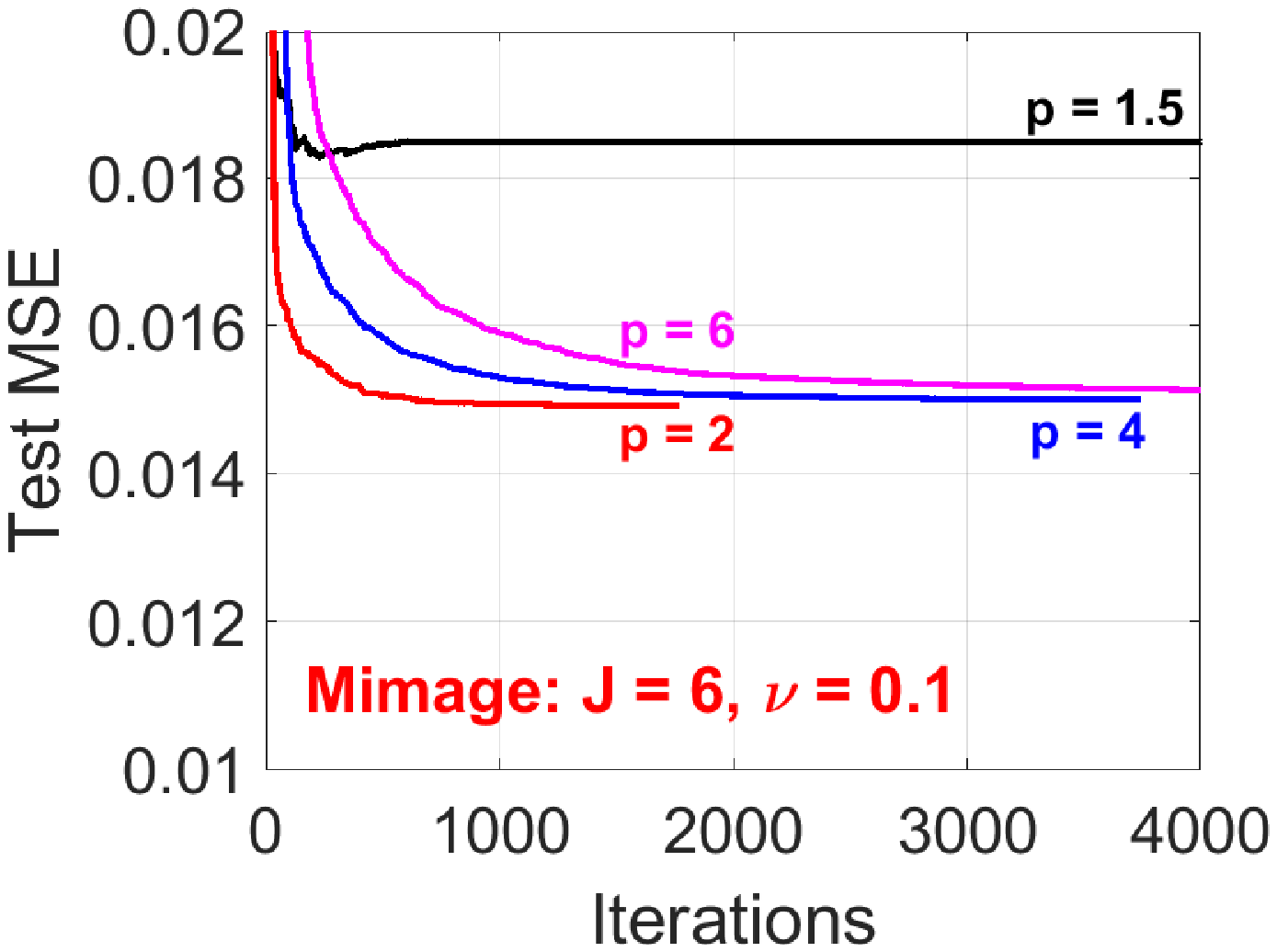}
    \includegraphics[width=2.2in]{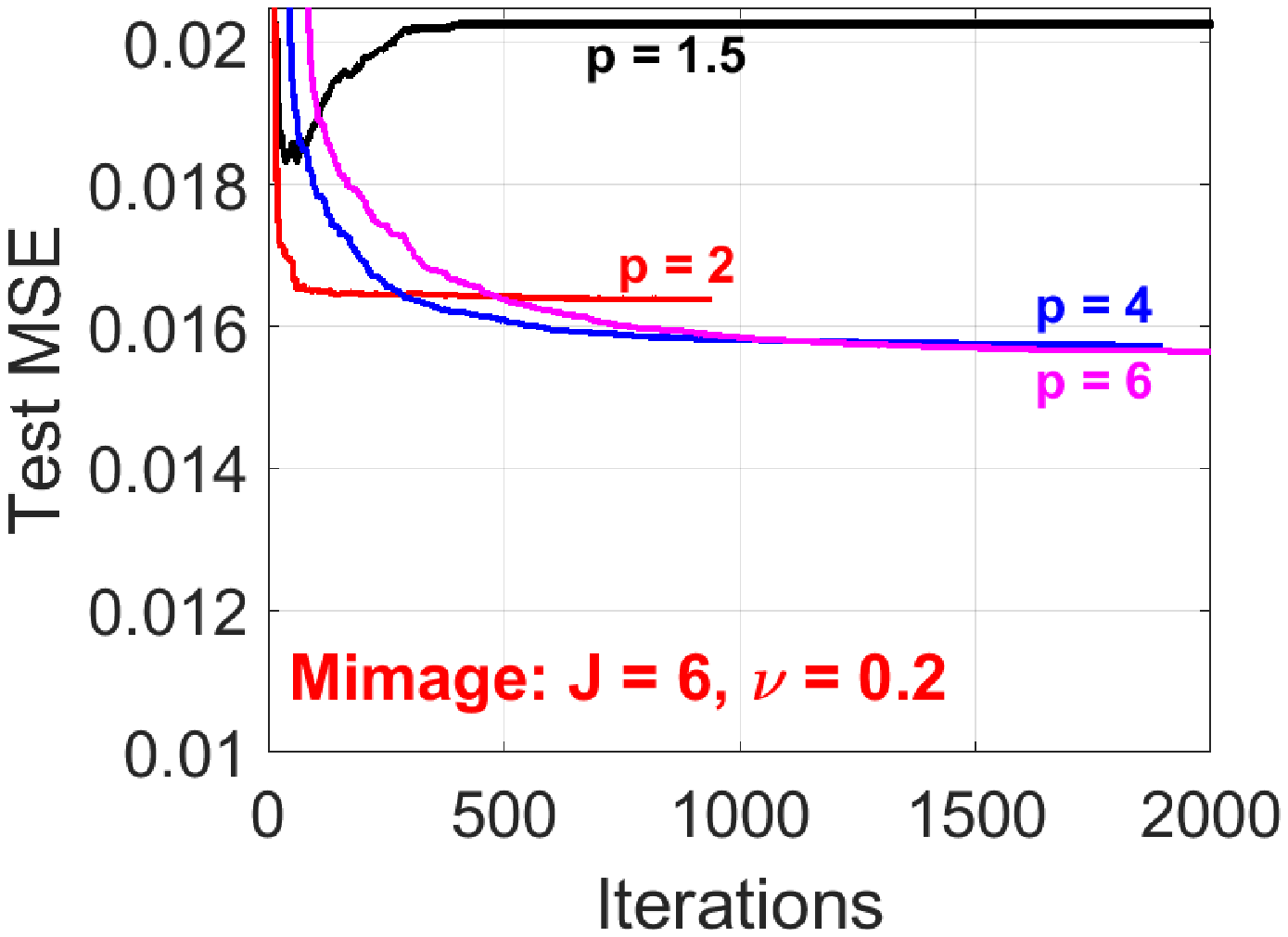}
}

\mbox{
    \includegraphics[width=2.2in]{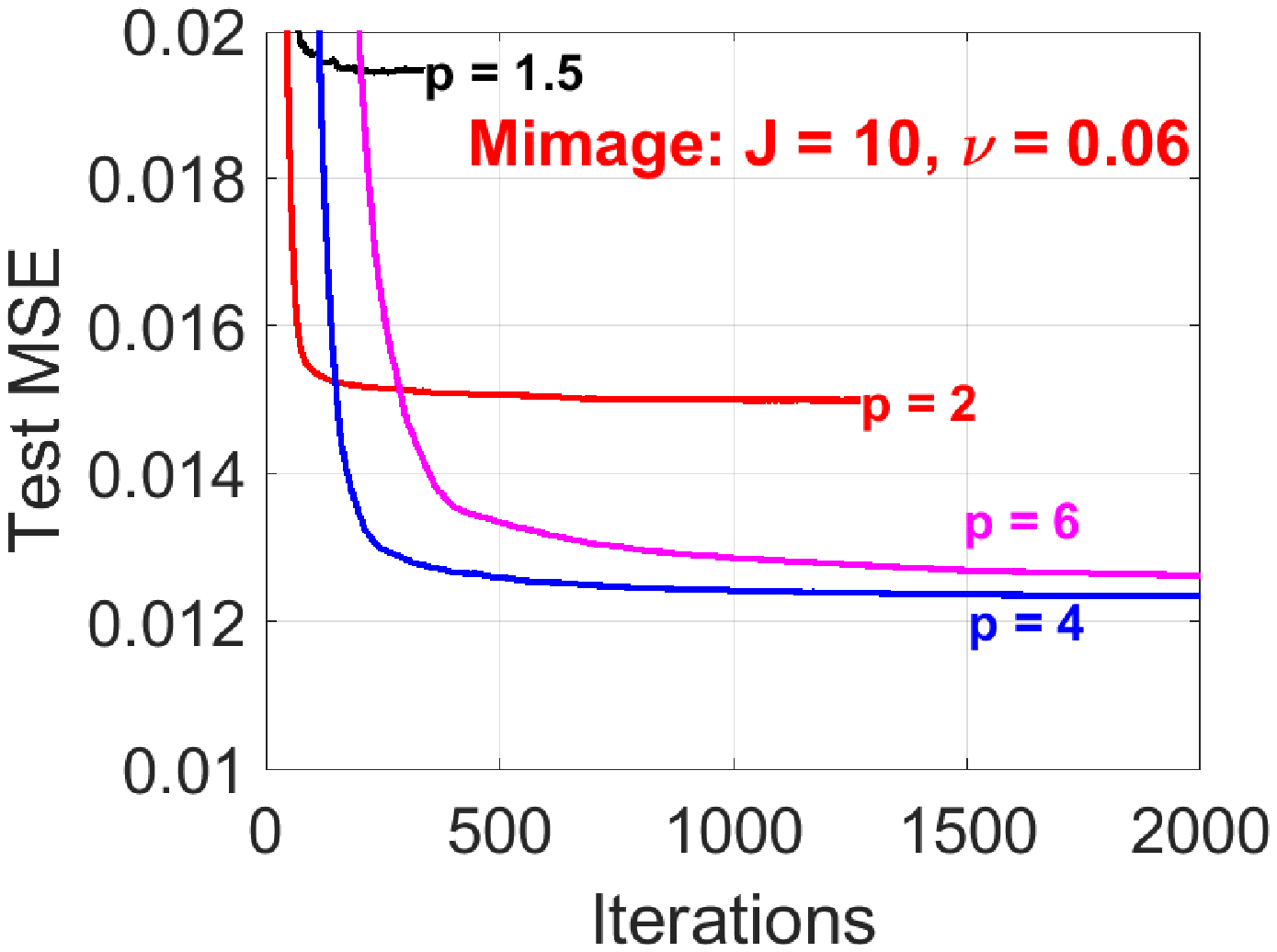}
    \includegraphics[width=2.2in]{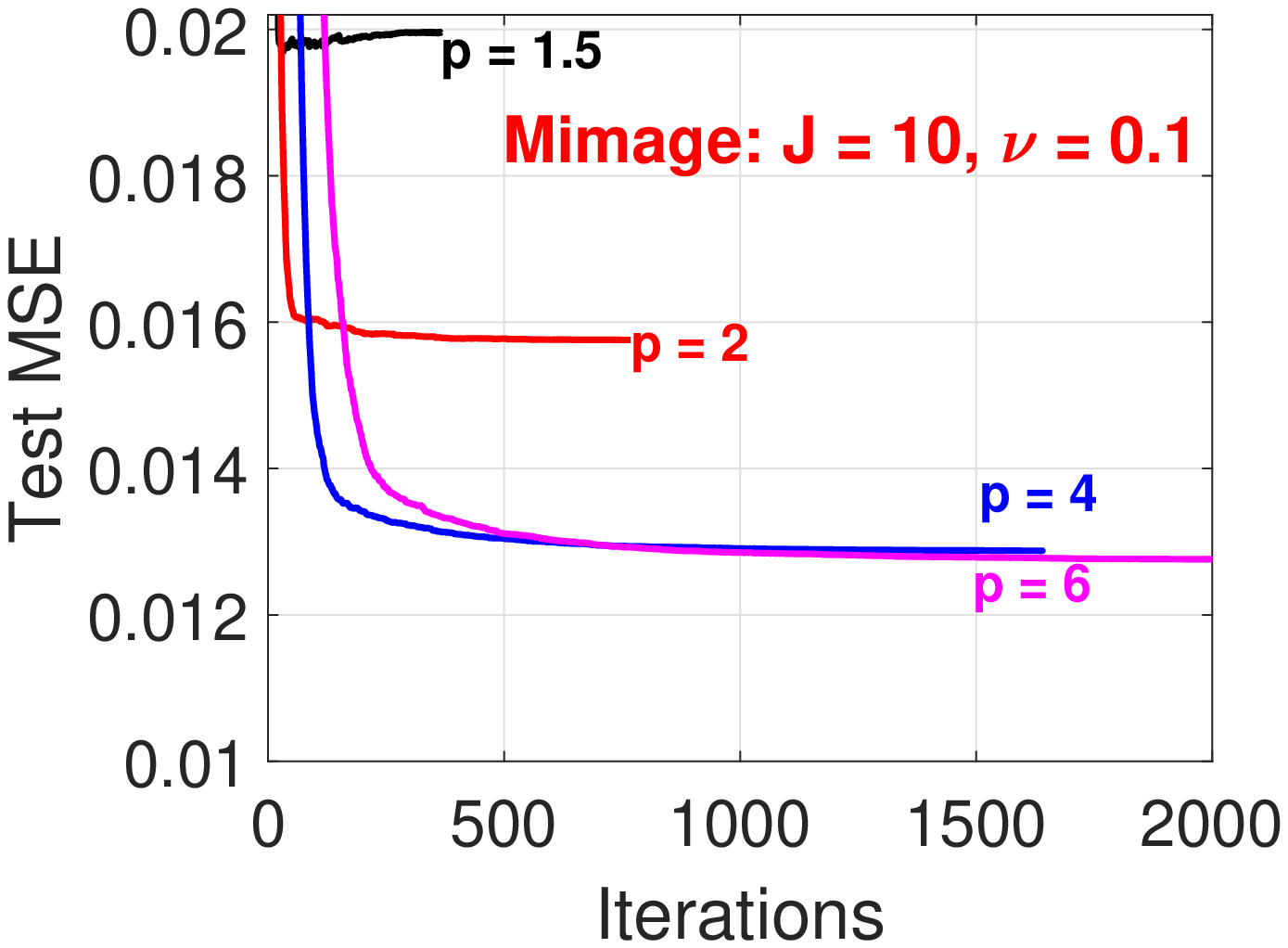}
    \includegraphics[width=2.2in]{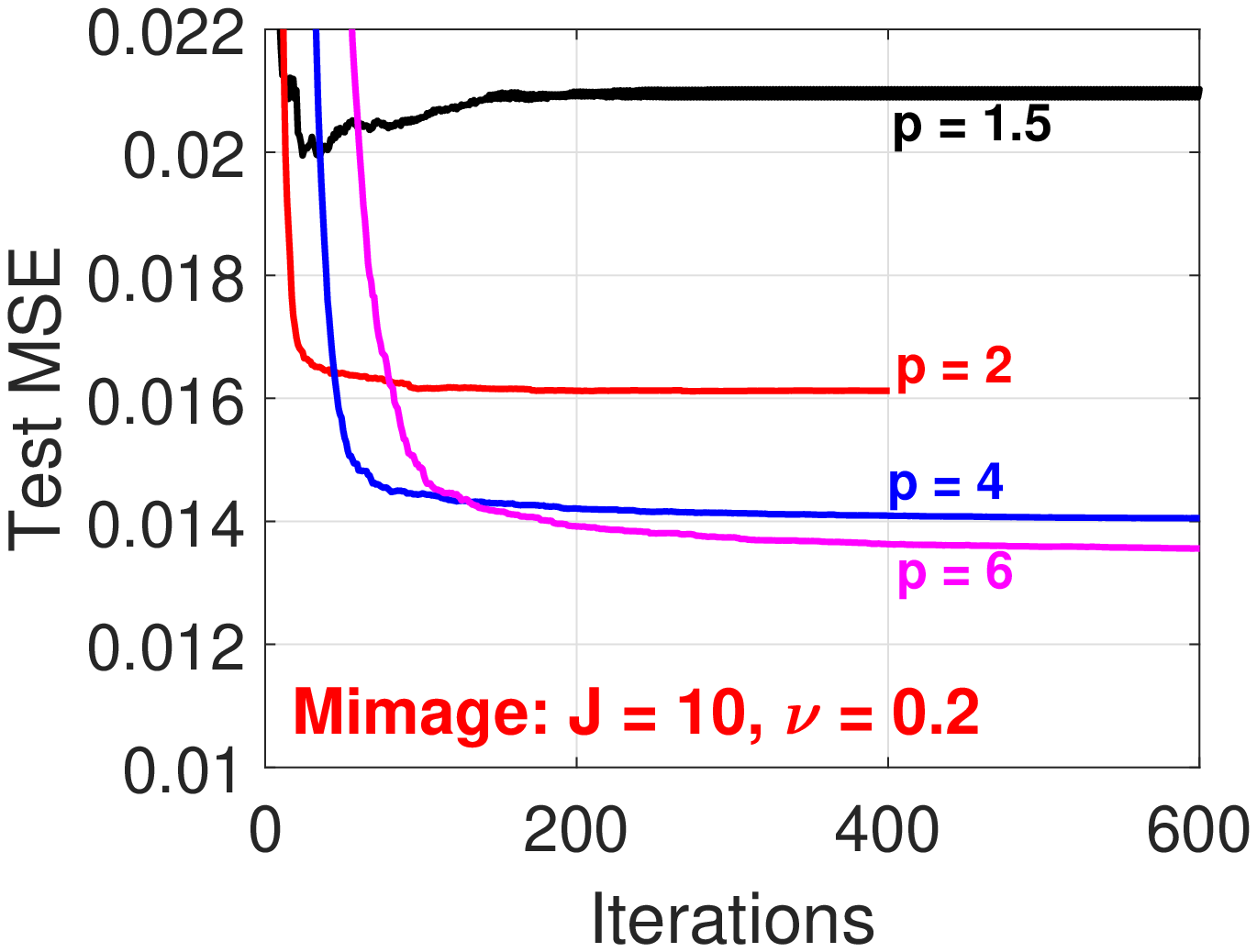}
}

\mbox{
    \includegraphics[width=2.2in]{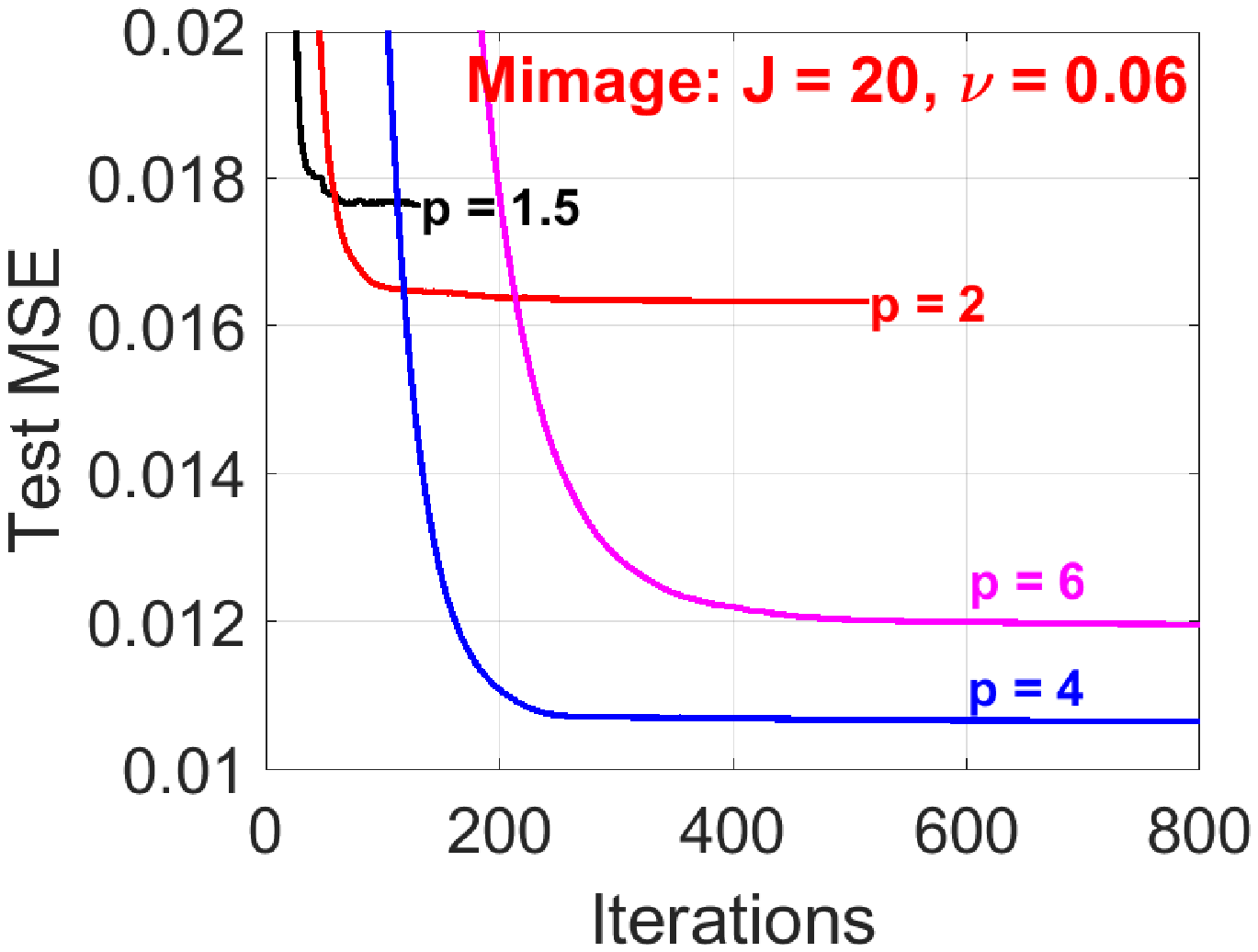}
    \includegraphics[width=2.2in]{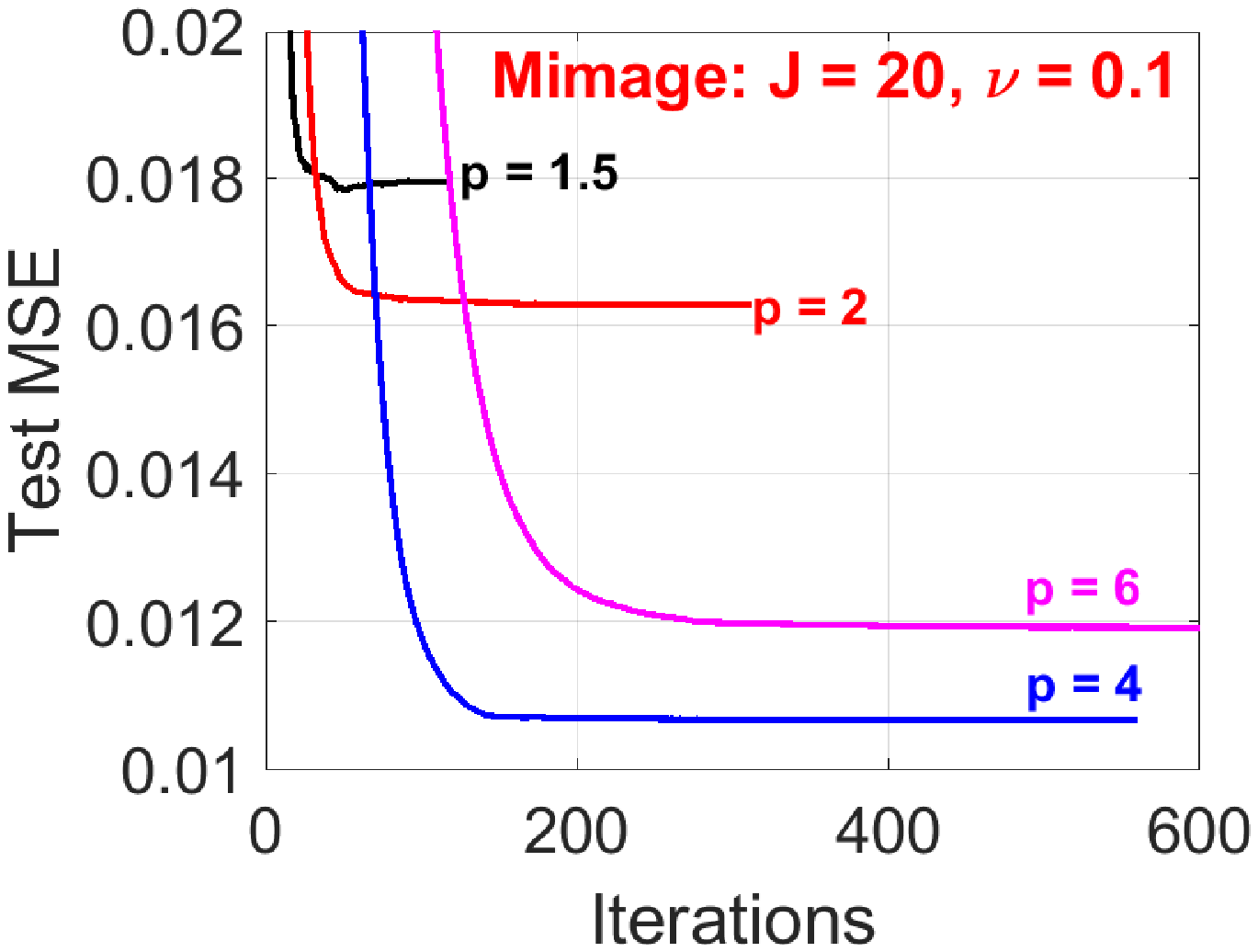}
    \includegraphics[width=2.2in]{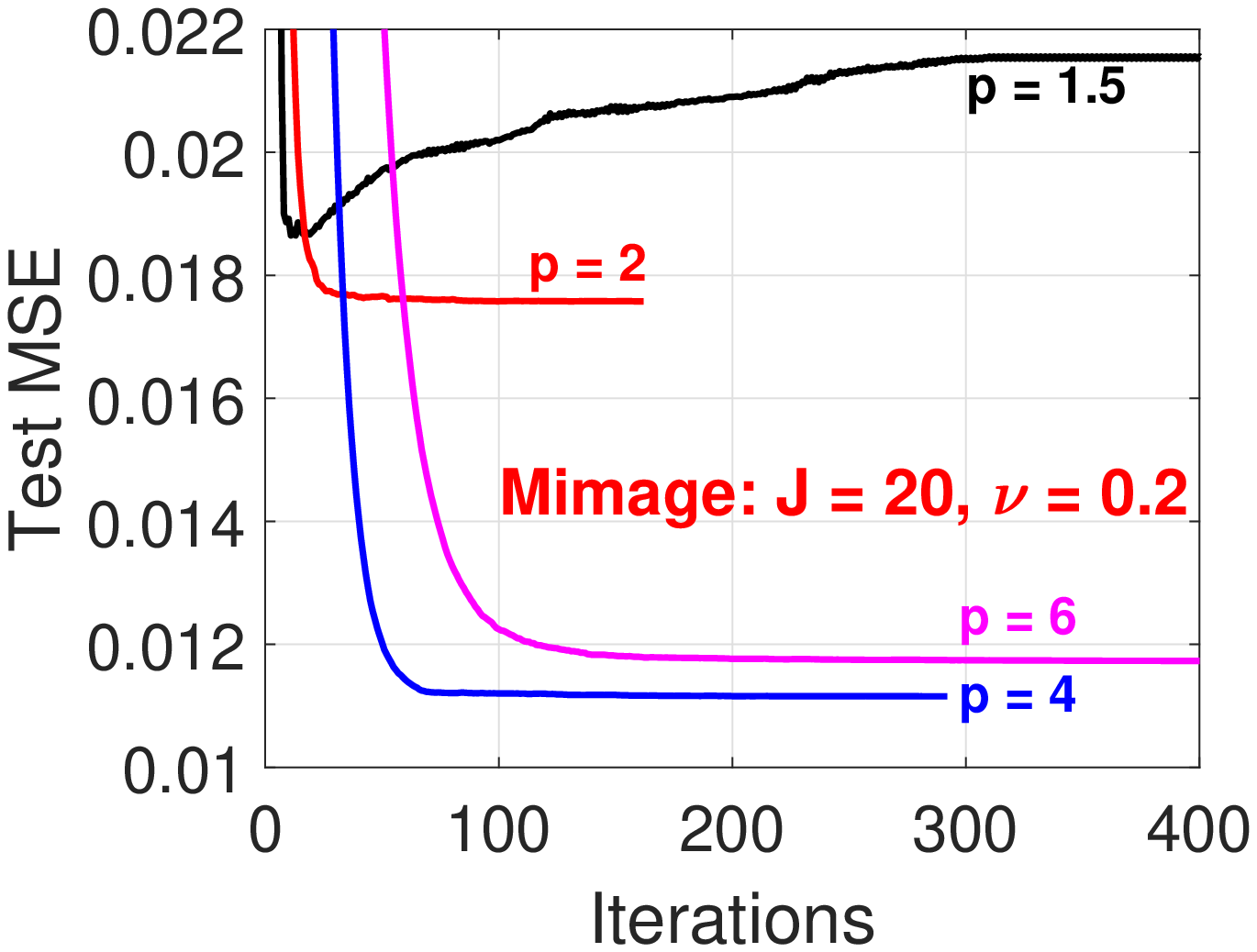}
}

\end{center}

    \caption{Test MSE history at each set of parameters $(J, \nu, p)$, for the Mimage dataset. The stopping criterion in Eq.~\eqref{eqn:stop} is used which appears to be perhaps too conservative for $p<2$.}
    \label{fig:MimageLpHistory}
\end{figure}

Figure~\ref{fig:MimageLpHistory} plots the  test $L_2$ losses (MSEs) for all the iterations of the Mimage dataset, at a particular set of parameters $J$, $\nu$, $p$. From the plots, it appears that the (conservative) stopping criterion in Eq.~\eqref{eqn:stop} is quite reasonable at least for $p\geq 2$. It might be too conservative for $p<2$.

\section{Conclusion}

In the context of regression, this paper combines two contributions: 1) kernel ridge regression using the pGMM kernel; and 2) $L_p$ boosting for regression. The simple (one-parameter only) pGMM kernel is surprisingly effective for regression. The parameter $p$ in the pGMM kernel is mathematically equivalent to applying a $p$-powered transformation on the original data before computing the GMM kernel. In practice, the pGMM kernel can be effectively hashed via consistent weighted sampling or extremal process sampling~\citep{li2021consistent}, which are numerically stable regardless of the $p$ values. If instead we directly apply the power transformation on the original data, numerical problems can easily occur. We hope the presented experiments on pGMM kernel for regression would help encourage the use of the pGMM kernel and variants in machine learning research and practice.

\vspace{0.1in}

\noindent $L_p$ boosting is also effective for regression. Different from the $p$ in the pGMM kernel, the parameter $p$ in $L_p$ boosting is for tuning the loss function used for training. While $L_2$ boosting is popular and standard in practice, using $L_p$ boosting is as convenient as $L_2$ when boosting is  equipped with the explicit gain formula using second derivatives of the loss function. Practitioners can simply treat the $p$ as an additional tuning parameter which might help achieve lower regression errors. $L_p$ boost has been implemented and included in the ``Fast ABC-Boost'' package~\citep{li2022package}. Typically, $L_p$ boosting could achieve lower regression errors than the pGMM kernel, but the gap in performance is often quite small especially if the original data are reasonably preprocessed.

\newpage\clearpage

\bibliographystyle{plainnat}
\bibliography{scholar_refs}

\end{document}